%% file: main.tex
\documentclass[letterpaper]{article} % DO NOT CHANGE THIS
\usepackage[preprint]{aaai2027}  % DO NOT CHANGE THIS
\usepackage[hyphens]{url}  % DO NOT CHANGE THIS
\usepackage{graphicx} % DO NOT CHANGE THIS
\usepackage{natbib}  % DO NOT CHANGE THIS AND DO NOT ADD ANY OPTIONS TO IT
\usepackage{caption} % DO NOT CHANGE THIS AND DO NOT ADD ANY OPTIONS TO IT
\usepackage{algorithm}
\usepackage{algorithmic}

\usepackage{newfloat}
\usepackage{listings}
\usepackage{amsmath}
\usepackage{amssymb}
\usepackage{multirow}
\usepackage{makecell}
\usepackage{array}
\usepackage{subcaption}
\usepackage{longtable}
\usepackage{flafter}
\newcolumntype{L}[1]{>{\raggedright\arraybackslash}p{#1}}
\newcolumntype{C}[1]{>{\centering\arraybackslash}p{#1}}
\DeclareCaptionStyle{ruled}{labelfont=normalfont,labelsep=colon,strut=off} % DO NOT CHANGE THIS
\floatstyle{ruled}
\newfloat{listing}{tb}{lst}{}
\floatname{listing}{Listing}

\usepackage{booktabs}

\title{ReCBM: Uncertainty-Gated Relational Reasoning for Concept Bottleneck Models}
\author {
    An Sui\textsuperscript{\rm 1},
    Yuzhu Li\textsuperscript{\rm 2},
    Fuping Wu\textsuperscript{\rm 3}\corresponding,
    Xiahai Zhuang\textsuperscript{\rm 1}\corresponding
}
\affiliations {
    \textsuperscript{\rm 1}School of Data Science, Fudan University, Shanghai, China\\
    \textsuperscript{\rm 2}Institute of Science and Technology for Brain-Inspired Intelligence, Fudan University, Shanghai, China.\\
    \textsuperscript{\rm 3}National Heart and Lung Institute, Imperial College London, London, United Kingdom.\\
    ansui23@m.fudan.edu.cn, 23110850033@m.fudan.edu.cn, f.wu1@imperial.ac.uk, zxh@fudan.edu.cn
}
\begin{document}

\maketitle

\begin{abstract}
Concept Bottleneck Models (CBMs) provide an interpretable framework by grounding predictions in human-understandable concepts, enabling semantic inspection and test-time intervention. Recent variants have improved CBMs through richer concept representations, uncertainty estimation, and dependency modeling. However, robust reasoning under unreliable concept states remains underexplored. Without such reasoning, misleading semantic evidence can propagate through the bottleneck, compromising both explanations and downstream predictions. To address this issue, we propose ReCBM, an uncertainty-gated relational reasoning framework for CBMs. ReCBM introduces semantically defined concept relations into the bottleneck and uses uncertainty to guide their refinement. By modeling co-occurrence, implication, and exclusion, ReCBM specifies how evidence is exchanged across concepts, while uncertainty modulates the contribution of each concept during this process. Experiments across diverse datasets showed that ReCBM improved concept and task recovery under missing and flipped concepts, supported uncertainty-aware intervention, and extracted compact task-relevant concept subsets without degrading downstream performance.
\end{abstract}


\section{Introduction}

Deep neural networks have substantially advanced a wide range of applications, yet their predictions are rarely grounded in explicit, human-verifiable evidence \citep{lecun2015deep,rudin,lipton2018mythos}. Concept Bottleneck Models (CBMs) address this limitation by routing predictions through human-interpretable concepts, making their decisions more transparent and editable \citep{CBM}. However, this interface is useful only when the exposed concept states are reliable. In practice, concept predictions can be noisy, missing, or inconsistent, while the human feedback used to correct them may be incomplete or uncertain \citep{park2025nips,intervention}. The resulting concept states may therefore remain unreliable, compromising both downstream predictions and the explanations presented to users. Robust CBMs should thus be able to use the reliable evidence that remains to recover concept states that are missing or identified as unreliable.

Recovering unreliable concept states requires determining which concepts can be trusted and using them to infer the remaining ones. Concept-wise uncertainty provides a reliability signal, while dependencies among concepts allow reliable states to support the recovery of missing or unreliable ones. For this process to remain interpretable, these dependencies should distinguish semantic relations such as co-occurrence, implication, and exclusion, which provide different forms of evidence for concept refinement \citep{deng2014large,li2015conditional,xiong2022hyperbolic}. Crucially, reliability and relational reasoning must operate jointly: uncertainty should regulate which concepts propagate evidence and which concepts receive correction. In interactive settings, the same mechanism should also accommodate reliability feedback when true concept values are unavailable. As summarized in Table~\ref{tab:capability_comparison}, existing CBM extensions provide uncertainty estimation or dependency modeling \citep{ProbCBM,SCBM,ECBM,GraphCBM}, but lack a unified mechanism for uncertainty-guided recovery through semantically defined relations.

To address this gap, we propose ReCBM, an uncertainty-gated relational reasoning framework that improves the robustness of CBMs. ReCBM combines concept relations with concept-wise uncertainty: reliable concepts provide relational evidence, while unreliable concepts are prevented from propagating errors and can be corrected using evidence from related concepts. An evidence-aware anchor preserves confident predictions when relational support is insufficient. This design enables interpretable concept recovery, while supporting uncertainty-aware intervention and compact concept subset extraction.

\begin{table}[!htbp]
    \centering
    \small
    \setlength{\tabcolsep}{2pt}
    \begin{tabular}{@{}l*{5}{w{c}{0.65cm}}@{}}
        \toprule
        Method
        & Un.
        & De.
        & Se.
        & U-ref.
        & U-int. \\
        \midrule
        CBM \citep{CBM}
        & $\times$ & $\times$ & $\times$ & $\times$ & $\times$ \\
        CEM \citep{CEM}
        & $\times$ & $\times$ & $\times$ & $\times$ & $\times$ \\
        ProbCBM \citep{ProbCBM}
        & $\checkmark$ & $\times$ & $\times$ & $\times$ & $\times$ \\
        SCBM \citep{SCBM}
        & $\checkmark$ & $\checkmark$ & $\times$ & $\times$ & $\triangle$ \\
        ECBM \citep{ECBM}
        & $\triangle$ & $\checkmark$ & $\times$ & $\times$ & $\times$ \\
        GraphCBM \citep{GraphCBM}
        & $\times$ & $\checkmark$ & $\times$ & $\times$ & $\times$ \\
        \midrule
        \textbf{ReCBM}
        & $\checkmark$ & $\checkmark$ & $\checkmark$ & $\checkmark$ & $\checkmark$ \\
        \bottomrule
    \end{tabular}
    \caption{Capabilities for recovering unreliable concepts, including concept uncertainty (Un.), concept dependencies (De.), semantic relations (Se.), uncertainty-gated refinement (U-ref.), and uncertainty-only intervention (U-int.). The symbols $\checkmark$, $\triangle$, and $\times$ indicate full, partial, and no explicit support, respectively.}
    \label{tab:capability_comparison}
\end{table}

We summarize our contributions as follows:
\begin{itemize}

\item We propose ReCBM, an uncertainty-gated relational reasoning framework that refines concept states using semantically defined co-occurrence, implication, and exclusion relations.

\item We introduce an uncertainty intervention mechanism that reduces the influence of uncertain concepts and corrects them using relational evidence, without direct concept intervention.

\item We develop a sufficient concept set extraction strategy that identifies compact global and class-specific subsets while preserving the predictive behavior of the full model through relational reconstruction.

\item Extensive experiments on diverse concept-based benchmarks demonstrated that ReCBM improved concept and task recovery under unreliable concept states, supported uncertainty-only intervention, and identified compact concept subsets while preserving task performance.

\end{itemize}

\section{Related Work}

Robust concept refinement requires interpretable concept states, reliability estimates, and mechanisms for exploiting concept dependencies.

\paragraph{Concept Bottleneck Models.}
Concept Bottleneck Models (CBMs) improve interpretability by decomposing prediction into concept prediction and label prediction, so that decisions are mediated by human-interpretable concepts \citep{CBM}. This structure enables concept-level explanations and test-time interventions, where users can edit predicted concepts to influence the final output. Subsequent work has extended CBMs along several directions. Concept Embedding Models (CEMs) improve expressiveness by representing each concept with a learnable embedding rather than a scalar variable \citep{CEM}. Post-hoc and label-free CBMs reduce the need for fully supervised concept annotations by constructing concept bottlenecks from pretrained models or vision-language representations \citep{PosthocCBM,LFCBM}. Other studies analyze limitations of CBMs, including concept leakage, weak concept alignment, and failures of local semantic grounding \citep{mahinpei2021promises,margeloiu2021concept,havasi2022addressing,raman2025locality}. Our work instead addresses recovery when the exposed concept state is
corrupted or unreliable.

\paragraph{Reliability and Uncertainty in Concept Prediction.}
Reliable concept estimates are essential for faithful CBM explanations and effective interventions. Standard CBMs usually produce deterministic concept predictions, which can be inadequate when concepts are ambiguous, noisy, or difficult to annotate. Probabilistic CBMs model concept uncertainty through probabilistic concept embeddings, allowing explanations to include both predictions and their uncertainty \citep{ProbCBM}. Other uncertainty-aware variants consider concept ambiguity, annotation uncertainty, or confidence-based intervention policies \citep{InteractiveCBM,eviCEM}. ReCBM is complementary to these approaches: it uses concept-wise uncertainty as an operational control variable inside the bottleneck, directly regulating which concepts may propagate evidence and which concepts should receive relational correction.

\paragraph{Concept Dependencies and Relational Reasoning.}
Many CBM formulations treat concepts as independent intermediate variables, although semantic concepts often exhibit strong dependencies. Stochastic CBMs capture correlations through a distributional parameterization and allow interventions on one concept to affect related concepts \citep{SCBM}. Energy-based CBMs define joint energy functions over inputs, concepts, and labels to model higher-order interactions \citep{ECBM}, while GraphCBM introduces graph-based concept dependencies \citep{GraphCBM}. ReCBM differs by organizing dependency propagation into co-occurrence, implication, and exclusion channels and coupling these channels with uncertainty-dependent source, receiver, and anchor gates. This design clarifies how concepts influence one another and how uncertainty regulates this process.

\section{Method}

\subsection{Framework Overview}
\label{subsec:framework_overview}
ReCBM refines unreliable concept states through uncertainty-weighted semantic relations before downstream prediction. Let $\mathcal{D}=\{(x_n,y_n,\mathbf{c}_n)\}_{n=1}^{N}$ denote the training set, where $x_n$ is the input, $y_n\in\{1,\ldots,K\}$ is the task label, and $\mathbf{c}_n\in\{0,1\}^{C}$ is a vector of $C$ predefined binary concepts.

\begin{figure*}[t]
    \centering
    \includegraphics[
        width=0.96\textwidth
    ]{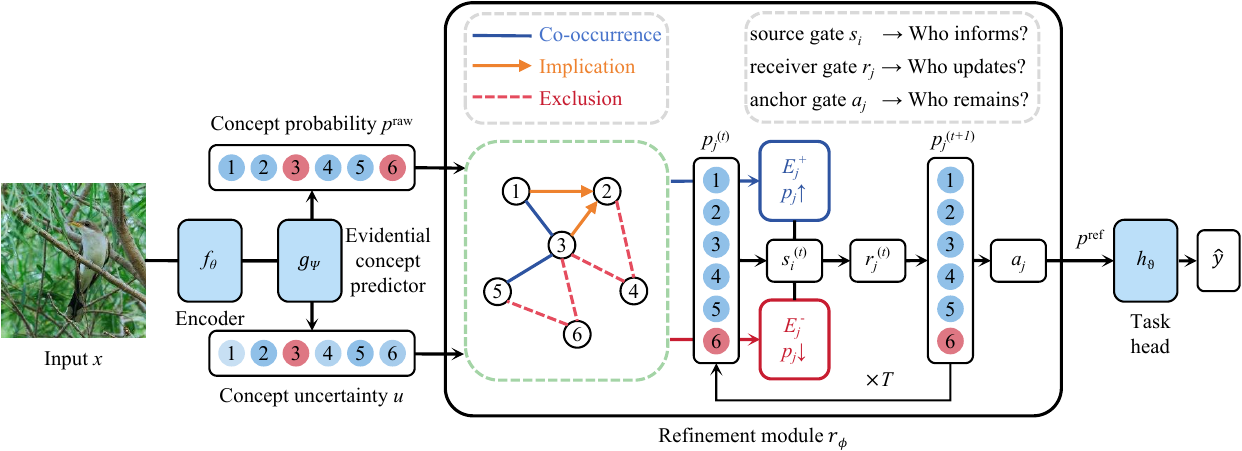}
    \caption{Overview of ReCBM. Numbered circles denote concepts, with red and blue representing relatively high and low values. For concept \(j\), \(E_j^{+}\) and \(E_j^{-}\) denote relational evidence that its probability should increase (\(\uparrow\)) or decrease (\(\downarrow\)), respectively.}
    \label{fig:overview}
\end{figure*}

As illustrated in Fig.~\ref{fig:overview}, the framework consists of four components. First, an encoder $f_{\theta}$ maps the input $x$ to a feature representation $\mathbf{z}=f_{\theta}(x)$. Second, an evidential concept predictor $g_{\psi}$ produces raw concept probabilities and concept-wise uncertainties, $(\mathbf{p}^{\mathrm{raw}},\mathbf{u})=g_{\psi}(\mathbf{z})$ with a Beta distribution \citep{EDL}. Third, an uncertainty-gated relational refinement module $r_{\phi}$ propagates evidence through semantic concept relations to obtain
$\mathbf{p}^{\mathrm{ref}}=r_{\phi}(\mathbf{p}^{\mathrm{raw}},\mathbf{u})$.
Finally, the task head produces the prediction
$\hat{y}=\arg\max_k
\left[h_{\vartheta}(\mathbf{p}^{\mathrm{ref}})\right]_k
$.

\subsection{Learning Concept Relations}
\label{subsec:relation_graph}

Reliable concept refinement requires distinguishing how concepts are related.
ReCBM therefore models co-occurrence, implication, and exclusion separately using three relation matrices
$\mathbf{A}^{\mathrm{co}}$, $\mathbf{A}^{\mathrm{imp}}$, and
$\mathbf{A}^{\mathrm{exc}}\in[0,1]^{C\times C}$, where larger matrix entries indicate stronger relations. Specifically, $A_{ij}^{\mathrm{co}}$ measures the tendency of concepts $i$ and $j$ to share the same state, $A_{ij}^{\mathrm{imp}}$ measures how strongly the presence of concept $i$ supports that of concept $j$, and $A_{ij}^{\mathrm{exc}}$ measures the tendency of concepts $i$ and $j$ not to be active simultaneously. Co-occurrence and exclusion are symmetric, whereas implication is directed. All diagonal entries are set to zero.

The relation matrices are initialized from pairwise concept statistics in the training set and subsequently optimized with the refinement module.
Symmetry is preserved for co-occurrence and exclusion during optimization. 
Details of the initialization and representative learned relations are provided in the supplementary material.

\subsection{Uncertainty-Gated Relational Refinement}
\label{subsec:relation_refinement}

Concept relations allow the state of one concept to provide evidence for another. However, a relational proposal is useful only when it is supported by sufficient evidence from reliable source concepts, and a target concept should accept this proposal only when correction is
warranted. ReCBM addresses these requirements through an iterative refinement process with receiver, source, and anchor gates.

\paragraph{Relational evidence and iterative refinement.}
For target concept \(j\), we aggregate positive and negative relational evidence from concept state \(\mathbf p\) using nonnegative source weights \(\mathbf w\):
\begin{align}
E_j^+(\mathbf w,\mathbf p)
={}&
\sum_i w_i p_i
\left(
A^{\mathrm{co}}_{ij}
+A^{\mathrm{imp}}_{ij}
\right),
\label{eq:relational_evidence1}
\\
E_j^-(\mathbf w,\mathbf p)
={}&
\sum_i w_i\Bigl[
A^{\mathrm{co}}_{ij}(1-p_i)
\notag\\
+&
A^{\mathrm{imp}}_{ji}(1-p_i)
+A^{\mathrm{exc}}_{ij}[p_i+p_j-1]_+
\Bigr].
\label{eq:relational_evidence2}
\end{align}
Here, \(E_j^+\) and \(E_j^-\) support the presence and absence of concept \(j\), respectively; \(w_i\) weights the contribution of source concept \(i\), and \([x]_+=\max(x,0)\).

Let $m_j(\mathbf w,\mathbf p)=E_j^+(\mathbf w,\mathbf p)+E_j^-(\mathbf w,\mathbf p)$ denote the total relational evidence. We define the relational proposal and its evidence-mass gate as
\begin{align}
q_j(\mathbf w,\mathbf p)
&=
\frac{E_j^+(\mathbf w,\mathbf p)}
{m_j(\mathbf w,\mathbf p)+\epsilon},
\label{eq:relation_pred1}
\\
g_j^{\mathrm{mass}}(\mathbf w,\mathbf p)
&=
\frac{m_j(\mathbf w,\mathbf p)}
{m_j(\mathbf w,\mathbf p)+1},
\label{eq:relation_pred2}
\end{align}
where \(\epsilon>0\) is a small constant for numerical stability. The proposal \(q_j\) specifies the state suggested by the related concepts, whereas the total evidence \(m_j\) controls how strongly this proposal influences the refinement through the evidence-mass gate \(g_j^{\mathrm{mass}}\). Specifically, \(g_j^{\mathrm{mass}}\) remains close to zero when the available relational evidence is weak and increases gradually toward one as the evidence mass grows.

Starting from \(\mathbf p^{(0)}=\mathbf p^{\mathrm{raw}}\), ReCBM performs \(T\) refinement iterations. At iteration \(t\), we use the \textit{source gate} \(\mathbf s^{(t)}\) defined later in Eq.~\eqref{eq:source_gate} as the source reliability weights, to compute
\begin{align}
q_j^{(t)}
&=
q_j(\mathbf s^{(t)},\mathbf p^{(t)}),
\\
g_j^{\mathrm{mass},(t)}
&=
g_j^{\mathrm{mass}}
(\mathbf s^{(t)},\mathbf p^{(t)}).
\end{align}
The target probability is then updated according to
\begin{equation}
p_j^{(t+1)}
=
\mathrm{clip}\!\left(
p_j^{(t)}
+
\delta
r_j^{(t)}
g_j^{\mathrm{mass},(t)}
\bigl(q_j^{(t)}-p_j^{(t)}\bigr),
\epsilon,1-\epsilon
\right),
\label{eq:iter_update}
\end{equation}
where \(\delta>0\) is a learned step size and \(r_j^{(t)}\) is the \textit{receiver gate} defined below.

\paragraph{Receiver gate.}
The receiver gate determines how strongly target concept \(j\) accepts the relational update:
\begin{equation}
r_j^{(t)}
=
\sigma\!\left(
\gamma_u u_j
+
\gamma_v
g_j^{\mathrm{mass},(t)}
\left|q_j^{(t)}-p_j^{(t)}\right|
+b_r
\right),
\label{eq:receiver_gate}
\end{equation}
where \(\sigma(\cdot)\) is the sigmoid function, \(\gamma_u,\gamma_v>0\) are learned coefficients, and \(b_r\) is a learned bias. A larger update is allowed when the current prediction is uncertain or when it differs substantially from a proposal supported by strong relational evidence.

\paragraph{Source gate.}
While the receiver gate controls the target's acceptance of an update, the source gate controls how strongly each source concept contributes to the relational proposal. A reliable source should have low uncertainty and agree with the estimate supported by its reliable neighbors.

To evaluate this consistency, we first compute an uncertainty-weighted preliminary proposal and its evidence mass:
\begin{align}
\bar q_i^{(t)}
&=
q_i(\mathbf 1-\mathbf u,\mathbf p^{(t)}),
\\
\bar g_i^{(t)}
&=
g_i^{\mathrm{mass}}
(\mathbf 1-\mathbf u,\mathbf p^{(t)}).
\end{align}
Thus, more certain concepts contribute more strongly to the preliminary proposal. We then measure the disagreement between concept \(i\) and this proposal:
\begin{equation}
v_i^{(t)}
=
\bar g_i^{(t)}
\left|
\bar q_i^{(t)}-p_i^{(t)}
\right|.
\end{equation}
The disagreement \(v_i^{(t)}\) becomes large only when concept \(i\)
differs from a well-supported preliminary proposal. Based on this
disagreement, the source gate is defined as
\begin{equation}
s_i^{(t)}
=
(1-u_i)
\sigma\!\left(
b_s-\gamma_s v_i^{(t)}
\right),
\label{eq:source_gate}
\end{equation}
where \(\gamma_s>0\) and \(b_s\) are learned parameters. The resulting \(\mathbf s^{(t)}\) is used as the source weights in the relational proposal.

\paragraph{Anchor gate.}
After \(T\) iterations, the anchor gate preserves reliable raw predictions that have weak initial relational support:
\begin{equation}
a_j
=
\left(1-g_j^{\mathrm{mass},(0)}\right)(1-u_j),
\label{eq:anchor}
\end{equation}
where a larger \(a_j\) assigns greater importance to the raw prediction.

Finally, the refined concept probability is obtained by:
\begin{equation}
p_j^{\mathrm{ref}}
=
a_jp_j^{\mathrm{raw}}
+
(1-a_j)p_j^{(T)}.
\end{equation}

\subsection{Training Objective}
\label{subsec:training_objective}

Relational refinement requires stable concept probabilities and informative uncertainty estimates. We therefore train ReCBM in two stages.

\paragraph{Stage 1: Evidential concept learning.}
In Stage~1, we jointly train the encoder \(f_\theta\), evidential concept predictor \(g_\psi\), and task predictor \(h_{\vartheta}\). Let $\Omega=\{(n,i):c_{n,i}\text{ is observed}\}$ index the available annotations, where $n$ and $i$ denote the sample and concept, respectively. For concept supervision, we use the evidential learning objective proposed in prior work~\citep{eviCEM,EDL}, which consists of an evidential negative log-likelihood \(\mathcal{L}_{\mathrm{concept}}\) and a KL regularizer \(\mathcal{L}_{\mathrm{KL}}\). We weight the latter by an annealing coefficient \(\omega_{\mathrm{KL}}\in[0,1]\) that increases linearly during the early stage of training. The task loss is \begin{equation} \mathcal{L}_{\mathrm{task}}=\mathrm{CE}\!\left(h_{\vartheta}(\mathbf{p}^{\mathrm{raw}}),y\right). \end{equation}
Here, $\mathrm{CE}$ denotes cross-entropy. Following the calibrated
uncertainty principle used in prior evidential models~\citep{zou2025devis},
we encourage incorrect concept predictions to have high uncertainty and
correct predictions to have low uncertainty. For each observed concept, we define the error indicator
$e_{n,i}=\mathbb{I}\!\left[
\mathbb{I}(p^{\mathrm{raw}}_{n,i}\geq 0.5)\neq c_{n,i}\right]$.

Let $\Omega_{\mathcal B}$ denote the set of observed sample--concept pairs in the current minibatch, and define the number of pairs with error label $\rho\in\{0,1\}$ as
$
\mathcal{N}(\rho)
=
\sum_{(n,i)\in\Omega_{\mathcal B}}
\mathbb{I}[e_{n,i}=\rho].
$
To prevent the more frequent error group from dominating training, we
use the group-balanced uncertainty loss
\begin{equation}
\mathcal{L}_{u}
=
\frac{
\sum_{(n,i)\in\Omega_{\mathcal B}}
\mathcal{N}(e_{n,i})^{-1}
\ell_{\mathrm{BCE}}(u_{n,i},e_{n,i})
}{
\sum_{(n,i)\in\Omega_{\mathcal B}}
\mathcal{N}(e_{n,i})^{-1}
},
\end{equation}
where $\ell_{\mathrm{BCE}}(u,e)=-e\log u-(1-e)\log(1-u).$

Finally, the Stage~1 objective is
\begin{align}
\mathcal{L}_{\mathrm{stage1}}
={}&
\lambda_c
\left(
\mathcal{L}_{\mathrm{concept}}
+\omega_{\mathrm{KL}}\mathcal{L}_{\mathrm{KL}}
\right)
\notag\\
&+
\lambda_y\mathcal{L}_{\mathrm{task}}
+\lambda_u\mathcal{L}_{u},
\label{eq:stage1_objective}
\end{align}
where $\lambda_c,\lambda_y,\lambda_u\geq0$ control the contributions of
concept supervision, task prediction, and uncertainty calibration,
respectively.

\paragraph{Stage 2: Relational refinement.}
During Stage~2, we freeze \(f_\theta\) and \(g_\psi\), and optimize the
refinement module \(r_\phi\) together with the task predictor
\(h_\vartheta\), initialized from Stage~1.

To expose \(r_\phi\) to unreliable concept states, we sample
\(\Omega_{\mathrm{aug}}\subseteq\Omega\) and replace each selected
probability with the value opposite to its label:
\begin{equation}
p_{n,i}^{\mathrm{raw,aug}}
=
\begin{cases}
0, & c_{n,i}=1,\\
1, & c_{n,i}=0,
\end{cases}
u_{n,i}^{\mathrm{aug}}=1,
(n,i)\in\Omega_{\mathrm{aug}}.
\end{equation}
All unselected entries retain their original probabilities and uncertainties. Let \(\mathbf p^{\mathrm{ref,aug}}\) denote the output obtained by refining the augmented state. We define the refinement loss as
\begin{align}
\mathcal{L}_{\mathrm{ref}}
={}&
\mathcal{B}(\mathbf p^{\mathrm{ref}},\mathbf c;\Omega)
+\kappa_{\mathrm{aug}}
\mathcal{B}(\mathbf p^{\mathrm{ref,aug}},\mathbf c;
\Omega_{\mathrm{aug}}),
\end{align}
where \(\mathbf c\) collects the ground-truth concepts,
\(\mathcal{B}(\mathbf p,\mathbf c;S)\) denotes binary cross-entropy over
an index set \(S\), averaged separately over its positive and negative
labels, and \(\kappa_{\mathrm{aug}}\geq0\) controls the strength of
augmentation supervision.

We further discourage refinement from moving an observed concept away
from its label. Let
\(\mathcal{D}_{\mathbf g}(\boldsymbol{\xi},\boldsymbol{\eta})\)
denote a gated directional penalty that measures whether an update from
\(\boldsymbol{\xi}\) to \(\boldsymbol{\eta}\) moves observed concept
probabilities away from their labels. We apply directional supervision at three points: to the final refined probabilities \(\mathbf p^{\mathrm{ref}}\), to the state \(\mathbf p^{(T)}\) produced by the \(T\) iterative relational updates, and to the relation-induced proposal \(\mathbf q^{(t)}\) at each iteration.  The resulting loss is
\begin{align}
\mathcal{L}_{\mathrm{dir}}
={}&
\tfrac{1}{2}
\mathcal{D}_{\mathbf g^0}
(\mathbf p^{\mathrm{raw}},\mathbf p^{\mathrm{ref}})
+\tfrac{1}{2}
\mathcal{D}_{\mathbf g^0}
(\mathbf p^{\mathrm{raw}},\mathbf p^{(T)})
\notag\\
&+
\frac{\kappa_{\mathrm{msg}}}{T}
\sum_{t=0}^{T-1}
\mathcal{D}_{\mathbf g^{\mathrm{mass},(t)}}
(\mathbf p^{(t)},\mathbf q^{(t)}),
\end{align}
where the first two terms supervise all observed entries and use the uniform gate \(\mathbf g^0\), defined by \(g^0_{n,i}=1\) for every \((n,i)\in\Omega\), and \(\kappa_{\mathrm{msg}}\geq0\) controls intermediate-message
supervision.

To define the gated directional penalty, let
\(\Omega_z=\{(n,i)\in\Omega:c_{n,i}=z\}\) for \(z\in\{0,1\}\), and let
\(\mathcal Z\subseteq\{0,1\}\) denote the label groups present in the current minibatch. We define
\begin{align}
&\mathcal{D}_{\mathbf g}
(\boldsymbol{\xi},\boldsymbol{\eta})
\notag\\
&=
\frac{1}{|\mathcal Z|}
\sum_{z\in\mathcal Z}
\frac{
\sum_{(n,i)\in\Omega_z}
g_{n,i}
\big[(2z-1)(\xi_{n,i}-\eta_{n,i})\big]_+
}{
\sum_{(n,i)\in\Omega_z}g_{n,i}+\epsilon
}.
\end{align}
Here, \(\boldsymbol{\xi}\) and \(\boldsymbol{\eta}\) denote the concept states before and after an update, respectively. The penalty is nonzero when an update decreases a positive-concept probability or increases a negative-concept probability. The gate \(g_{n,i}\) weights the directional penalty for each entry, while averaging over the nonempty label groups prevents either group from dominating the loss.

The primary Stage~2 objective is
\begin{align}
\mathcal{L}_{\mathrm{stage2}}
={}&
\lambda_y
\mathrm{CE}\!\left(h_\vartheta(\mathbf p^{\mathrm{ref}}),y\right)
+\lambda_{\mathrm{ref}}\mathcal{L}_{\mathrm{ref}}
+\lambda_{\mathrm{dir}}\mathcal{L}_{\mathrm{dir}},
\end{align}
where \(\lambda_{\mathrm{ref}}\) and
\(\lambda_{\mathrm{dir}}\) are nonnegative loss weights. Auxiliary
optimization terms are detailed in the supplementary material.

\subsection{Sufficient Concept Set Extraction}
\label{subsec:minimal_sets}

Concept relations may allow part of the bottleneck to be reconstructed from the remaining concepts. We therefore seek compact concept subsets that preserve both concept reconstruction and downstream prediction. To identify such subsets, we define an anchor-weighted task-gradient importance score for each concept and use this score to guide a greedy search.

Specifically, the importance of concept \(i\) is defined as
\begin{equation}
\mathrm{score}_i =
\left(
\frac{1}{N_{\mathrm{val}}}
\sum_{n=1}^{N_{\mathrm{val}}}
\left|
\frac{\partial \ell_{n,y_n}}
{\partial p^{\mathrm{ref}}_{n,i}}
\right|
\right)
\left(
\frac{1}{N_{\mathrm{val}}}
\sum_{n=1}^{N_{\mathrm{val}}}
a_{n,i}
\right),
\end{equation}
where \(N_{\mathrm{val}}\) is the number of validation samples, \(\ell_{n,y_n}\) is the logit of the ground-truth class, and \(a_{n,i}\) is the preservation anchor defined in Eq.~\eqref{eq:anchor}. The first factor measures the sensitivity of the task prediction to concept \(i\). The second factor measures how strongly the refined prediction retains the corresponding raw concept value. Their product therefore prioritizes concepts that are important to the task and cannot be readily reconstructed from the remaining concepts.

Given a candidate set \(S\), we construct the refinement input as
\begin{equation}
(\tilde p_i,\tilde u_i)=
\begin{cases}
(p_i^{\mathrm{raw}},0), & i\in S,\\
(0.5,1), & i\notin S.
\end{cases}
\end{equation}
Thus, selected concepts retain their raw predictions, while unselected concepts are treated as unknown. The relation module reconstructs \(\tilde{\mathbf p}^{\mathrm{ref}}\) from this partial concept state. We define concept and task retention as the validation concept F1 and task accuracy obtained from \(\tilde{\mathbf p}^{\mathrm{ref}}\), respectively,
divided by their full-model counterparts obtained from \(\mathbf p^{\mathrm{ref}}\). A set \(S\) is considered sufficient if both
retention values are at least \(100\%\). We rank concepts by \(\mathrm{score}_i\) and greedily search the resulting order for a compact set satisfying both conditions. Full search and pruning details are provided in the supplementary material.

Repeating the search for each class \(y\) yields a class-specific set \(S_y\). At test time, the concepts in \(S_y\) are supplied with their ground-truth values, while the remaining concepts are treated as unknown. We evaluate each \(S_y\) as a one-vs-rest classifier, treating \(y\) as positive and all other classes as negative, and report balanced accuracy averaged over the classes for which a sufficient set is found.

\section{Experiments}
\label{sec:experiments}

Our experiments assessed whether relational refinement recovered unreliable concepts, which components drove this recovery, and whether uncertainty supported intervention and compact concept-set extraction.

\subsection{Experimental Setup}
\label{subsec:experimental_setup}

\subsubsection{Datasets}
\label{subsubsec:datasets}

We evaluated ReCBM on two image datasets with human-annotated concepts and a controlled synthetic dataset with predefined concept relations. Table~\ref{tab:dataset_statistics} summarizes their statistics.

\begin{table}[!htbp]
    \centering
    \small
    \setlength{\tabcolsep}{2pt}
    \begin{tabular}{ccccc}
        \toprule
        Dataset & Train & Validation & Test & Concepts / Classes \\
        \midrule
        {WBC}       & 6,169  & 1,030 & 3,099 & 24 / 5 \\
        {CUB}   & 4,796  & 1,198 & 5,794 & 112 / 200 \\
        {Synthetic} & 12,000 & 3,000 & 3,000 & 12 / 4 \\
        \bottomrule
    \end{tabular}
    \caption{Dataset statistics.}
    \label{tab:dataset_statistics}
\end{table}

\textit{WBC.}
The White Blood Cell Attribute dataset (\textit{WBC}) contains peripheral blood cell images from five leukocyte classes~\citep{WBCDataset}. We converted its 11 categorical morphological attributes into 24 binary concepts.

\textit{CUB.}
The Caltech-UCSD Birds-200-2011 dataset contains 11,788 images from 200 bird species~\citep{CUBDataset}. Following standard CBM preprocessing~\citep{CBM}, we retained the 112 binary attributes as concepts.

\textit{Synthetic.}
We constructed a controlled dataset with 12 binary concepts, 4 classes, and predefined co-occurrence, implication, and exclusion relations. Full generation details are provided in the supplementary material.

\subsubsection{Baselines}
\label{subsubsec:baselines}

We compared ReCBM with five baseline configurations: independently and jointly trained CBMs \citep{CBM}, ProbCBM \citep{ProbCBM}, SCBM \citep{SCBM}, and GraphCBM \citep{GraphCBM}.

\subsubsection{Implementation Details}
\label{subsubsec:implementation_details}

For \textit{WBC} and \textit{CUB}, we used a ResNet-34 concept encoder with images resized to \(224\times224\). For \textit{Synthetic}, we used an MLP with hidden dimension 128. ReCBM was trained for up to 150 epochs, with 70 epochs for Stage~1
and 80 epochs for Stage~2. The initial learning rate was
\(5\times10^{-4}\) for \textit{WBC} and \textit{CUB} and \(10^{-3}\) for \textit{Synthetic}.

The relation matrices were initialized from training-set concept statistics and optimized during Stage~2. We used \(T=5\) refinement iterations for \textit{WBC} and \textit{Synthetic} and \(T=10\) for \textit{CUB}. For Stage~2 augmentation, each observed concept entry was included in
\(\Omega_{\mathrm{aug}}\) with probability \(0.1\). In Stage~1, we set
\(\lambda_c=\lambda_y=1\) and \(\lambda_u=10\), and linearly annealed
\(\omega_{\mathrm{KL}}\) from 0 to 1 over the first 10 epochs. In Stage~2,
we set \(\lambda_y=1\), \(\lambda_{\mathrm{dir}}=10\), and
\(\kappa_{\mathrm{msg}}=1\). We used
\((\lambda_{\mathrm{ref}},\kappa_{\mathrm{aug}})=(20,0.25)\) for
\textit{WBC} and \((10,1)\) for \textit{CUB} and \textit{Synthetic}.
All experiments were conducted on a single NVIDIA GeForce RTX 3090 GPU.
Additional training hyperparameters are provided in the supplementary material.

\begin{figure}[t]
    \centering
    \includegraphics[width=\columnwidth]{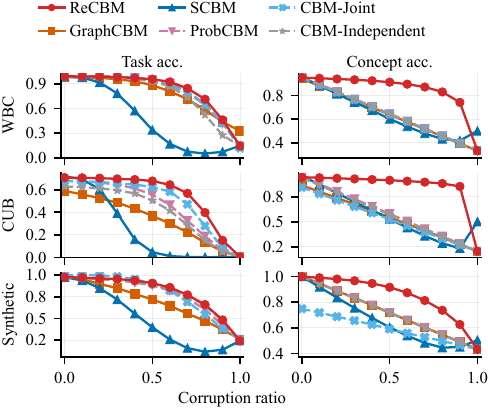}
    \caption{Recovery from missing concepts. Concept and task accuracy are reported as the proportion of entries assigned the neutral state increases.}
    \label{fig:native_corruption_missing}
\end{figure}

\begin{figure}[t]
    \centering
    \includegraphics[width=\columnwidth]{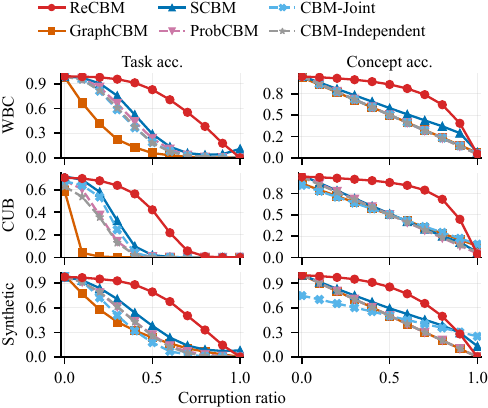}
    \caption{Recovery from concept flips. Concept and task accuracy are reported as the proportion of flipped entries assigned maximal uncertainty increases.}
    \label{fig:native_corruption_flip}
\end{figure}

\subsection{Robustness to Concept Corruption}
\label{subsec:corruption_robustness}

We evaluated the robustness of ReCBM at evenly spaced corruption
ratios from 0 to 1, defined as \(r_k = k/10\) for indices \(k\) ranging
from 0 to 10.
At test time, selected concept entries were replaced by
\((p^{\mathrm{raw}}_{n,i},u_{n,i})=(0.5,1)\) for \textit{missing}
corruption, or by
\((p^{\mathrm{raw}}_{n,i},u_{n,i})=(1-\hat c_{n,i},1)\) for
\textit{flip} corruption, where
\(\hat c_{n,i}=\mathbb{I}[p^{\mathrm{raw}}_{n,i}\geq0.5]\).
% We evaluated corruption ratios \(r_k=k/K\), where \(K=10\), and \(r=0\) denoted evaluation without corruption. 
Robustness across all ratios was summarized by the trapezoidal Corruption Robustness AUC:
\begin{equation}
    \mathrm{CR\text{-}AUC}
    =\sum_{k=0}^{K-1}
    \frac{\mathrm{Acc}(r_k)+\mathrm{Acc}(r_{k+1})}{2}
    (r_{k+1}-r_k).
\end{equation}
Higher CR-AUC indicates better robustness across corruption severities.

% We excluded CEM and ECBM because their concept embeddings or joint energy inference could retain input-dependent information. Details of the corruption protocol applied to the remaining methods are provided in the supplementary material.

\begin{table}[!t]
    \centering
    \small
    \setlength{\tabcolsep}{2pt}
    \begin{tabular}{cccccc}
        \toprule
        \multirow{2}{*}{Method} & \multicolumn{2}{c}{Missing} & \multicolumn{2}{c}{Flip} & Macro \\
        \cmidrule(lr){2-3}\cmidrule(lr){4-5}
        & Concept & Task & Concept & Task & CR-AUC \\
        \midrule
        GraphCBM & 63.20 & 60.62 & 50.13 & 18.61 & 48.14 \\
        SCBM & 61.08 & 38.49 & 56.38 & 36.06 & 48.00 \\
        ProbCBM & 64.70 & 68.80 & 50.11 & 30.10 & 53.43 \\
        CBM-Joint & 58.62 & 70.51 & 50.13 & 28.59 & 51.96 \\
        CBM-Independent & 64.49 & 66.53 & 50.11 & 28.66 & 52.45 \\
        \midrule
        \textbf{ReCBM} & \textbf{86.59} & \textbf{73.72} & \textbf{76.01} & \textbf{57.64} & \textbf{73.49} \\
        \bottomrule
    \end{tabular}
    \caption{CR-AUC (\%) averaged across datasets. Macro averages missing/flip and concept/task CR-AUC.}
    \label{tab:corruption_robustness_auc}
\end{table}

Table~\ref{tab:corruption_robustness_auc} shows that ReCBM achieved the highest concept and task CR-AUC under both corruption types, demonstrating recovery from both missing and misleading concept evidence.

Figure~\ref{fig:native_corruption_missing} shows that ReCBM degraded more slowly as concepts were replaced by neutral states. Relational refinement reconstructed missing states from the remaining reliable concepts, delaying the loss of both concept and task information. This advantage diminished near full corruption, where little reliable evidence remained for reconstruction.

Figure~\ref{fig:native_corruption_flip} presents a harder setting in which corrupted concepts provided active but incorrect evidence. ReCBM limited their influence through the source gate and admitted corrections supported by reliable neighbors, yielding substantially slower degradation across most corruption ratios. SCBM's rebound at full corruption reflected its fallback to the training prior when no reliable concepts remained, rather than concept recovery.

The starting points of Figures~\ref{fig:native_corruption_missing} and~\ref{fig:native_corruption_flip} correspond to performance without
corruption (\(r=0\)). ReCBM remained competitive with existing methods
in this setting, indicating that its robustness gains did not come at
the cost of substantially degraded performance without corruption.

\subsection{Ablation Study}
\label{subsec:ablation_analysis}

We ablated each relation type individually and jointly, as well as adaptive gating. For the latter, the source and receiver gates were fixed to one and the preservation anchor to zero, yielding fully ungated relational updates.
The ablation was conducted on \textit{WBC} under $50\%$ missing and flip corruption. Results on the remaining datasets are provided in the supplementary material.

\begin{table}[!htbp]
    \centering
    \small
    \setlength{\tabcolsep}{2pt}
    \begin{tabular}{@{}cccc|cc|cc@{}}
        \toprule
        \multirow{2}{*}{C}
        & \multirow{2}{*}{I}
        & \multirow{2}{*}{E}
        & \multirow{2}{*}{G}
        & \multicolumn{2}{c|}{Flip}
        & \multicolumn{2}{c}{Missing} \\
        \cmidrule(lr){5-6}\cmidrule(l){7-8}
        & & & & Concept & Task & Concept & Task \\
        \midrule
        \checkmark & \checkmark & \checkmark & \checkmark
        & \textbf{80.10$\pm$3.14}
        & \textbf{76.00$\pm$2.62}
        & \textbf{90.89$\pm$0.21}
        & \textbf{95.31$\pm$0.49} \\

        $\times$ & \checkmark & \checkmark & \checkmark
        & 74.53$\pm$1.70
        & 59.75$\pm$1.42
        & 90.61$\pm$0.07
        & 94.87$\pm$0.26 \\

        \checkmark & $\times$ & \checkmark & \checkmark
        & 72.89$\pm$0.21
        & 71.81$\pm$0.50
        & 88.70$\pm$0.15
        & 94.84$\pm$0.69 \\

        \checkmark & \checkmark & $\times$ & \checkmark
        & 78.09$\pm$5.03
        & 65.87$\pm$7.34
        & 89.23$\pm$0.36
        & 94.85$\pm$0.53 \\

        \checkmark & $\times$ & $\times$ & \checkmark
        & 64.53$\pm$1.46
        & 55.02$\pm$1.12
        & 86.42$\pm$0.14
        & 93.97$\pm$0.40 \\

        $\times$ & \checkmark & $\times$ & \checkmark
        & 59.24$\pm$0.97
        & 30.54$\pm$1.81
        & 85.69$\pm$0.05
        & 93.99$\pm$0.47 \\

        $\times$ & $\times$ & \checkmark & \checkmark
        & 66.02$\pm$0.25
        & 50.31$\pm$1.05
        & 81.20$\pm$0.07
        & 94.30$\pm$0.41 \\

        $\times$ & $\times$ & $\times$ & \checkmark
        & 49.94$\pm$0.11
        & 13.92$\pm$0.63
        & 64.44$\pm$0.14
        & 91.84$\pm$0.86 \\

        \checkmark & \checkmark & \checkmark & $\times$
        & 74.83$\pm$0.06
        & 14.71$\pm$0.99
        & 85.77$\pm$0.12
        & 90.59$\pm$1.85 \\
        \bottomrule
    \end{tabular}%
    \caption{Component ablation on \textit{WBC} under 50\% concept corruption. C, I, E, and G denote co-occurrence, implication, exclusion, and adaptive gating. Concept and task accuracy (\%) are reported as mean \(\pm\) standard deviation over three paired seeds. Best means are in bold.}
    \label{tab:ablation_corruption}
\end{table}

As presented in Table \ref{tab:ablation_corruption}, the complete model obtained the highest mean concept and task accuracy under both corruption settings. The sharp drop in task accuracy without adaptive gating, particularly under flips, showed that controlling unreliable propagation was critical when corrupted concepts provided active evidence.

The relation ablations further showed that co-occurrence, implication, and exclusion provided complementary recovery signals. No individual or pairwise configuration matched the complete model across both corruption types, while removing all relation channels caused the largest loss in concept recovery. Compared to \textit{missing} setting, the substantially larger task degradation under \textit{flips} further indicated that relational evidence was most important when the bottleneck contained actively misleading states.

\subsection{Uncertainty Intervention}
\label{subsec:uncertainty_intervention}

We evaluated whether identifying unreliable concepts enabled relational recovery without providing their true values. We first corrupted concept predictions using a sampling ratio of \(0.5\) by replacing each sampled binary prediction with its opposite value. We then marked an increasing
fraction of the corrupted entries as unreliable. For each marked entry, its uncertainty was set to \(u=1\), while its corrupted probability remained unchanged. Recovery therefore relied on the remaining reliable concepts and their relations.

\begin{figure}[t]
    \centering
    \includegraphics[width=\columnwidth]
    {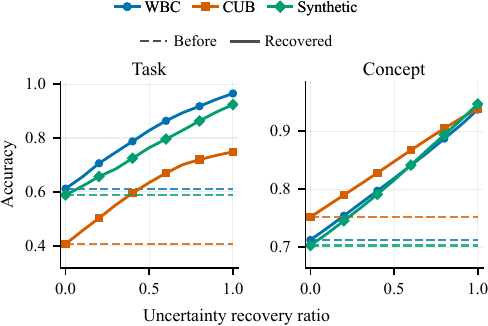}
    \caption{Uncertainty-guided recovery on \textit{WBC}, \textit{CUB}, and \textit{Synthetic}. Dashed lines denote performance before recovery, and solid lines denote performance as increasing fractions of corrupted entries were identified as unreliable.}
    \label{fig:uncertainty_recovery}
\end{figure}

Figure~\ref{fig:uncertainty_recovery} shows that both concept and task accuracy improved as more corrupted entries were identified as unreliable, demonstrating that reliability annotations alone could support relational recovery. This interface is useful when an annotator or monitoring system can detect an unreliable concept more readily than determine its correct value. Further experiments in the supplementary material examine how the assigned uncertainty controls the influence of provided concept edits.

\subsection{Sufficient Concept Set Extraction}
\label{subsec:sufficient_concept_sets}

We evaluated whether compact concept subsets could preserve the concept and task performance of ReCBM. Following the extraction procedure in Section~\ref{subsec:minimal_sets}, selection was performed exclusively on the validation set, yielding one global set shared by all classes and a separate set for each class. Each selected set was subsequently evaluated once on the test set. For class-specific evaluation, only the selected concepts were supplied with their ground-truth values.

Table~\ref{tab:sufficient_concept_sets} summarizes the global and class-specific extraction results. For the global set, Size gives the number of selected concepts relative to the full concept vocabulary, and Task-Ret gives its test task accuracy as a percentage of that achieved by the full ReCBM. For the class-specific sets, Valid reports the number of classes for which the selected subset retained at least \(100\%\) of both the full-model concept F1 and task accuracy on the validation set. Mean-Size and BAcc report, respectively, the average subset size and macro one-vs-rest balanced accuracy over these valid classes.

\begin{table}[!htbp]
    \centering
    \small
    \setlength{\tabcolsep}{2pt}
    \begin{tabular}{@{}cccccc@{}}
        \toprule
        \multirow{2}{*}{Dataset}
        & \multicolumn{2}{c}{Global Set}
        & \multicolumn{3}{c}{Class-Specific Sets} \\
        \cmidrule(lr){2-3}\cmidrule(l){4-6}
        & Size & Task-Ret & Valid & Mean-Size & BAcc \\
        \midrule
        \textit{WBC}
        & $18/24$ & $99.90\%$ & $5/5$ & $6.60/24$ & $98.67\%$ \\
        \textit{CUB}
        & $99/112$ & $101.03\%$ & $166/200$ & $8.42/112$ & $97.38\%$ \\
        \textit{Synthetic}
        & $12/12$ & $100.03\%$ & $4/4$ & $10.50/12$ & $98.56\%$ \\
        \bottomrule
    \end{tabular}
    \caption{Global and class-specific concept sets selected on the
    validation set and evaluated on the test set.}
    \label{tab:sufficient_concept_sets}
\end{table}

Table~\ref{tab:sufficient_concept_sets} reveals global concept redundancy on \textit{WBC} and \textit{CUB}, where relational refinement reconstructs unselected concepts without reducing task performance. In contrast, \textit{Synthetic} requires the complete concept set, indicating less global redundancy.

Class-specific extraction produces substantially smaller sets while achieving high one-vs-rest balanced accuracy, indicating that individual classes can be supported by compact, class-dependent concept subsets.

\section{Conclusion}
\label{sec:conclusion}

In this work, we introduced ReCBM, an uncertainty-gated relational reasoning framework that refines concept predictions through semantically defined relations. By controlling evidence propagation according to concept reliability, ReCBM improved concept and task robustness under missing and flipped concepts. It also supported uncertainty-only recovery without ground-truth concept values and identified compact class-specific concept subsets. Future work will explore iterative human-in-the-loop refinement and relations derived from external knowledge.

\bibliography{aaai2027}

\clearpage
\onecolumn
\begin{center}
{\Large\bfseries Supplementary Material}
\end{center}
\input{supplementary}

\end{document}

%% file: supplementary.tex
\section*{Contents}
\begingroup
\setlength{\parindent}{0pt}
\setlength{\parskip}{3pt}
\textbf{Guide to the Supplementary Material}
\dotfill \pageref{app:guide}\par
\textbf{Appendix A: Synthetic Dataset Details}
\dotfill \pageref{app:synthetic_generation}\par
\textbf{Appendix B: Relation-Matrix Initialization}
\dotfill \pageref{app:relation_initialization}\par
\textbf{Appendix C: Preserving Relation Semantics during Optimization}
\dotfill \pageref{app:auxiliary_optimization}\par
\textbf{Appendix D: Implementation Details}
\dotfill \pageref{app:implementation_details}\par
\textbf{Appendix E: ReCBM Training Pseudocode}
\dotfill \pageref{app:pseudocode}\par
\textbf{Appendix F: Complete Ablation Results}
\dotfill \pageref{app:complete_ablation}\par
\textbf{Appendix G: Qualitative Refinement Analysis}
\dotfill \pageref{subsec:qualitative_analysis}\par
\textbf{Appendix H: Uncertainty-Only Intervention}
\dotfill \pageref{app:uncertainty_recovery}\par
\textbf{Appendix I: Counterfactual Edits under Different Uncertainty Levels}
\dotfill \pageref{app:uncertainty_counterfactual}\par
\textbf{Appendix J: Sufficient Concept-Set Extraction}
\dotfill \pageref{app:concept_set_extraction}\par
\endgroup
\FloatBarrier

\appendix

\section*{Guide to the Supplementary Material}
\label{app:guide}

The supplementary material is organized as follows.  Appendix~A specifies the complete Synthetic dataset generator.  Appendix~B describes relation-matrix initialization and presents representative learned relations. Appendix~C describes how the semantic consistency of the relation matrices is preserved during optimization.  Appendix~D reports implementation details, including training, corruption, hyperparameters for ReCBM and the comparison methods, random seeds, and the computing environment.  Appendix~E gives the complete ReCBM training pseudocode.  Appendix~F reports ablation results on all datasets.  Appendix~G provides qualitative examples of concept refinement and uncertainty-only intervention.  Appendix~H describes the uncertainty-only intervention protocol.  Appendix~I studies how assigned uncertainty controls counterfactual concept edits.  Finally, Appendix J describes the complete procedure for extracting and pruning sufficient concept sets, together with the resulting class-specific sets.

\clearpage

\section{Synthetic Dataset Details}
\label{app:synthetic_generation}

This section documents the Synthetic dataset construction to clarify how its
concepts, task labels, and relational structure were generated.
The Synthetic dataset represents access-control events with 12 binary concepts and 4 task labels. Table~\ref{tab:synthetic_vocabulary} defines the complete concept and task-label vocabularies. The concepts describe authentication outcome, device and location status, failed-login behavior, account privilege, and policy enforcement. The task labels summarize four event types: normal access, credential anomaly, device or location anomaly, and policy block or privilege risk.

\begin{table*}[htbp]
    \centering
    \small
    \begin{tabular}{@{}C{0.14\textwidth}C{0.68\textwidth}@{}}
        \toprule
        Index & Concept \\
        \midrule
        0 & Password verified \\
        1 & Multi-Factor Authentication (MFA) verified \\
        2 & Registered device \\
        3 & New device \\
        4 & Usual location \\
        5 & Anomalous location \\
        6 & Failed attempts reach threshold \\
        7 & Account locked \\
        8 & Login succeeds \\
        9 & Administrator account \\
        10 & Administrator operation \\
        11 & Login blocked by policy \\
        \midrule
        Index & Task label \\
        \midrule
        0 & Normal access \\
        1 & Credential anomaly \\
        2 & Device or location anomaly \\
        3 & Policy block or privilege risk \\
        \bottomrule
    \end{tabular}
    \caption{Concept and task-label vocabularies of the Synthetic dataset.}
    \label{tab:synthetic_vocabulary}
\end{table*}

We sampled task labels uniformly and drew concepts independently from label-dependent Bernoulli distributions. Table~\ref{tab:synthetic_base_probabilities} reports the complete probability vector in the order of the concept indices. We then imposed the relations in Table~\ref{tab:synthetic_graph}: for each co-occurring pair, activation of one concept activated the other with probability $0.85$; implications were enforced by activating the consequent whenever the antecedent was active; and exclusion conflicts were resolved using the class-dependent rules described below. This procedure produced a concept vector consistent with the specified relational structure.

\begin{table*}[htbp]
    \centering
    \small
    \setlength{\tabcolsep}{2pt}
    \begin{tabular}{@{}C{0.20\textwidth}*{12}{C{0.052\textwidth}}@{}}
        \toprule
        Label & $c_0$ & $c_1$ & $c_2$ & $c_3$ & $c_4$ & $c_5$ &
        $c_6$ & $c_7$ & $c_8$ & $c_9$ & $c_{10}$ & $c_{11}$ \\
        \midrule
        Normal access
        &.98&.96&.92&.02&.90&.02&.01&.01&.94&.06&.02&.01\\
        Credential anomaly
        &.35&.08&.48&.08&.44&.08&.96&.94&.01&.03&.02&.92\\
        Device/location anomaly
        &.92&.88&.02&.94&.02&.92&.04&.02&.78&.03&.02&.03\\
        Policy block/privilege risk
        &.62&.56&.16&.18&.14&.16&.04&.03&.01&.96&.90&.82\\
        \bottomrule
    \end{tabular}
    \caption{Label-conditional Bernoulli probabilities used before applying
    the relations. Concept indices follow
    Table~\ref{tab:synthetic_vocabulary}.}
    \label{tab:synthetic_base_probabilities}
\end{table*}

After independently sampling the 12 initial concept states from Table~\ref{tab:synthetic_base_probabilities}, the generator applied the relations in the following order.
\begin{enumerate}
    \item For each co-occurrence pair, if exactly one concept was active, the other was activated with probability \(0.85\).
    
    \item All implication rules were applied repeatedly until no implication could activate an additional concept. For example, an active ``Failed attempts reach threshold'' concept activated ``Account locked,'' which subsequently activated ``Login blocked by policy.''

    \item For each exclusion pair, mutual exclusivity was enforced by retaining one active concept according to a class-conditional selection rule. If both ``Registered device'' and ``New device'' were active, the latter was retained for the device/location-anomaly class. For either of the other non-normal classes, it was retained with probability \(0.35\); otherwise, ``Registered device'' was retained. The same rule was used for ``Usual location'' and ``Anomalous location.'' Independently of the class, an active ``Account locked'' or ``Login blocked by policy'' concept deactivated ``Login succeeds.''

    \item Implication closure and exclusion resolution were applied once more because resolving one relation could change whether another relation was satisfied. The resulting vector was used as the final concept annotation.
\end{enumerate}

The training, validation, and test splits contained 12,000, 3,000, and 3,000 examples, respectively. The generator mapped each final concept vector to a 32-dimensional input in two sampling stages. First, it sampled one projection matrix and four class-specific offset vectors:
\begin{align}
\mathbf W_{dk} &\overset{\mathrm{i.i.d.}}{\sim}\mathcal N(0,1),
&& d\in\{1,\ldots,32\},\quad k\in\{1,\ldots,12\}, \\
\mathbf e_y &\overset{\mathrm{i.i.d.}}{\sim}
\mathcal N(\mathbf 0,\sigma_e^2\mathbf I),
&& y\in\{0,1,2,3\},
\end{align}
where \(\mathbf W\in\mathbb R^{32\times12}\), \(\mathbf e_y\in\mathbb R^{32}\), and \(\mathbf I\in\mathbb R^{32\times32}\) is the identity matrix. We set \(\sigma_e=0.35\). The sampled \(\mathbf W\) and the four vectors \(\{\mathbf e_y\}_{y=0}^{3}\) were fixed and shared by the training, validation, and test splits.

Second, for every example \(n\), the generator independently sampled
\[
\boldsymbol\epsilon_n
\sim
\mathcal N(\mathbf 0,\sigma_\epsilon^2\mathbf I),
\qquad
\sigma_\epsilon=0.35,
\]
and constructed the unstandardized input
\begin{equation}
\widetilde{\mathbf x}_n
=
\mathbf W\mathbf c_n+\mathbf e_{y_n}+\boldsymbol\epsilon_n.
\label{eq:supp_synthetic_input}
\end{equation}
Here, \(\mathbf c_n\in\{0,1\}^{12}\) and \(y_n\in\{0,1,2,3\}\) are the concept vector and task label of example \(n\), respectively. The term \(\mathbf W\mathbf c_n\) encodes the concept state, \(\mathbf e_{y_n}\) adds the offset shared by all examples of class \(y_n\), and \(\boldsymbol\epsilon_n\) introduces example-specific variation.

Finally, each split \(s\in\{\mathrm{train},\mathrm{validation},\mathrm{test}\}\) was standardized independently in every feature dimension:
\begin{equation}
x_{n,d}
=
\frac{\widetilde{x}_{n,d}-\mu_{s,d}}
{\sigma_{s,d}+10^{-6}},
\end{equation}
where \(\mu_{s,d}\) and \(\sigma_{s,d}\) are the empirical mean and standard deviation of dimension \(d\) in split \(s\). All random quantities were generated using a NumPy random generator initialized with seed 42.

\begin{table*}[htbp]
    \centering
    \small
    \begin{tabular}{@{}C{0.22\textwidth}C{0.68\textwidth}@{}}
        \toprule
        Relation & Concept pairs \\
        \midrule
        \multirow{5}{*}{Co-occurrence}
        & Password verified $\leftrightarrow$ MFA verified \\
        & Registered device $\leftrightarrow$ Usual location \\
        & New device $\leftrightarrow$ Anomalous location \\
        & Failed attempts reach threshold $\leftrightarrow$ Account locked \\
        & Administrator account $\leftrightarrow$ Administrator operation \\
        \midrule
        \multirow{5}{*}{Implication}
        & Login succeeds $\rightarrow$ Password verified \\
        & Login succeeds $\rightarrow$ MFA verified \\
        & Failed attempts reach threshold $\rightarrow$ Account locked \\
        & Account locked $\rightarrow$ Login blocked by policy \\
        & Administrator operation $\rightarrow$ Administrator account \\
        \midrule
        \multirow{4}{*}{Exclusion}
        & Registered device $\perp$ New device \\
        & Usual location $\perp$ Anomalous location \\
        & Account locked $\perp$ Login succeeds \\
        & Login blocked by policy $\perp$ Login succeeds \\
        \bottomrule
    \end{tabular}
    \caption{Ground-truth relations in the Synthetic dataset.}
    \label{tab:synthetic_graph}
\end{table*}

\clearpage

\section{Relation-Matrix Initialization}
\label{app:relation_initialization}

This section describes the complete initialization procedure. For a pair of binary concepts $(c_i,c_j)$, let $\hat{\nu}_{ab}^{ij}=n_{ab}^{ij}/N_{ij}$ denote the empirical frequency of $(c_i,c_j)=(a,b)$ among the $N_{ij}$ training examples in which both concepts are observed, where \(a,b\in\{0,1\}\), \(n_{ab}^{ij}\) is the corresponding joint count, and \(i,j\in\{1,\ldots,C\}\) index concepts. Under independence, the expected frequencies are
\begin{equation}
\nu_{11}^{ij}=p_i p_j,\qquad
\nu_{10}^{ij}=p_i(1-p_j),\qquad
\nu_{01}^{ij}=(1-p_i)p_j,
\end{equation}
where $p_i=P(c_i=1)$ and $p_j=P(c_j=1)$. The normalized reduction in violation frequency and its
expected-support factor are
\begin{equation}
\rho_{ab}^{ij}=
\left[
\frac{\nu_{ab}^{ij}-\mathrm{UCB}(\hat{\nu}_{ab}^{ij})}
{\max(\nu_{ab}^{ij},\epsilon)}
\right]_+,
\qquad
R_{ab}^{ij}=
\frac{N_{ij}\nu_{ab}^{ij}}
{N_{ij}\nu_{ab}^{ij}+\tau+\epsilon}.
\label{eq:supp_reduction}
\end{equation}
For an empirical frequency $\hat{\nu}=n/N$, the Wilson upper confidence
bound~\citep{Wilson} used in our implementation is
\begin{equation}
\mathrm{UCB}(\hat{\nu})
=
\frac{
\hat{\nu}+\frac{z^2}{2N}
+z\sqrt{\frac{\hat{\nu}(1-\hat{\nu})}{N}+\frac{z^2}{4N^2}}
}{
1+\frac{z^2}{N}
}.
\label{eq:supp_wilson}
\end{equation}
We used $z=1.96$, corresponding to a two-sided 95\% interval, and set the
support constant \(\tau=5\) and numerical constant
\(\epsilon=10^{-6}\). The operator $[x]_+=\max(x,0)$. The initial relation strengths are
\begin{align}
A_{ij}^{\mathrm{co}} &={}
\sqrt{\rho_{10}^{ij}\rho_{01}^{ij}R_{10}^{ij}R_{01}^{ij}},
\notag\\
A_{ij}^{\mathrm{imp}} &={}
[\rho_{10}^{ij}-\rho_{01}^{ij}]_+R_{10}^{ij},
\notag\\
A_{ij}^{\mathrm{exc}} &={}\rho_{11}^{ij}R_{11}^{ij}.
\label{eq:supp_relation_init}
\end{align}
Co-occurrence and exclusion were symmetrized by averaging each matrix with its transpose. All diagonal entries were set to zero, and all strengths were clipped to $[0,1]$. Co-occurrence required both one-sided mismatch cells to be rarer than independence predicted, implication required the forward violation cell to be selectively rare, and exclusion required joint activation to be rare. These values initialized trainable relation parameters. The matrices were subsequently optimized with the rest of the refinement module.

\paragraph{Representative Learned Relations}
\mbox{}\par
Table~\ref{tab:supp_learned_relations} lists the two learned relations with the highest weights for each relation type and dataset.

\begin{table*}[htbp]
    \centering
    \small
    \begin{tabular}{|C{0.13\textwidth}|C{0.14\textwidth}|C{0.66\textwidth}|}
        \hline
        Dataset & Type & Strongest learned relations \\
        \hline
        \multirow[c]{6}{*}{WBC}
        & \multirow[c]{2}{*}{Co-occurrence}
        & round granule type $\leftrightarrow$ red granule colour (0.993) \\
        & & nil granule type $\leftrightarrow$ nil granule colour (0.991) \\
        \cline{2-3}
        & \multirow[c]{2}{*}{Implication}
        & coarse granule type $\rightarrow$ granularity (0.928) \\
        & & segmented multilobed nucleus shape $\rightarrow$ granularity (0.928) \\
        \cline{2-3}
        & \multirow[c]{2}{*}{Exclusion}
        & nil granule type $\perp$ granularity (0.991) \\
        & & nil granule colour $\perp$ granularity (0.991) \\
        \hline
        \multirow[c]{6}{*}{CUB}
        & \multirow[c]{2}{*}{Co-occurrence}
        & belly color: yellow $\leftrightarrow$ primary color: yellow
          (0.971) \\
        & & underparts color: yellow $\leftrightarrow$ primary color:
          yellow (0.971) \\
        \cline{2-3}
        & \multirow[c]{2}{*}{Implication}
        & forehead color: yellow $\rightarrow$ bill length: shorter than head
          (0.919) \\
        & & forehead color: yellow $\rightarrow$ shape: perching-like
          (0.913) \\
        \cline{2-3}
        & \multirow[c]{2}{*}{Exclusion}
        & bill length: about the same as head $\perp$ bill length: shorter than
          head (0.994) \\
        & & size: small $\perp$ size: medium (0.987) \\
        \hline
        \multirow[c]{6}{*}{Synthetic}
        & \multirow[c]{2}{*}{Co-occurrence}
        & Failed attempts reach threshold $\leftrightarrow$ Account locked
          (0.993) \\
        & & Administrator account $\leftrightarrow$ Administrator
          operation (0.987) \\
        \cline{2-3}
        & \multirow[c]{2}{*}{Implication}
        & Login succeeds $\rightarrow$ Password verified (0.915) \\
        & & Login succeeds $\rightarrow$ MFA verified (0.909) \\
        \cline{2-3}
        & \multirow[c]{2}{*}{Exclusion}
        & Login succeeds $\perp$ Login blocked by policy (0.998) \\
        & & Registered device $\perp$ New device (0.997) \\
        \hline
    \end{tabular}
    \caption{Highest-weight learned concept relations. Values in parentheses are the corresponding learned relation weights.}
    \label{tab:supp_learned_relations}
\end{table*}

\clearpage

\section{Preserving Relation Semantics during Optimization}
\label{app:auxiliary_optimization}

To preserve the intended distinction among co-occurrence, implication, and
exclusion during optimization, we used a logic-conflict regularizer. Let $C$ be the number of concepts and let $\mathbf A^{\mathrm{co}},\mathbf A^{\mathrm{imp}}, \mathbf A^{\mathrm{exc}}\in[0,1]^{C\times C}$ denote the three relation matrices defined in the main paper.  For matrices $\mathbf X,\mathbf Y$, $\mathbf X\odot\mathbf Y$ denotes element-wise multiplication, $\mathbf X^\top$ denotes transpose, and \(\langle\mathbf X\rangle=C^{-2}\sum_{i=1}^{C}\sum_{j=1}^{C}X_{ij}\) denotes the mean of all entries.  We write \(\mathbf 1\in\mathbb R^{C\times C}\) for the all-ones matrix.  The regularizer is
\begin{align}
\mathcal L_{\mathrm{logic}}
={}&
\left\langle\mathbf A^{\mathrm{co}}\odot
\mathbf A^{\mathrm{exc}}\right\rangle
+\left\langle
(\mathbf A^{\mathrm{imp}}+\mathbf A^{\mathrm{imp}\top})
\odot\mathbf A^{\mathrm{exc}}\right\rangle
\notag\\
&+\left\langle
(\mathbf A^{\mathrm{imp}}\odot\mathbf A^{\mathrm{imp}\top})
\odot(\mathbf 1-\mathbf A^{\mathrm{co}})
\right\rangle .
\label{eq:supp_logic_loss}
\end{align}
The first term penalizes overlap between co-occurrence and exclusion for the same concept pair. The second penalizes overlap between either direction of an implication and exclusion. The third maps strong bidirectional implication to co-occurrence by penalizing disagreement between the two relation types.

Stage~1 optimized \(h_\vartheta\) using raw concept probabilities. At the beginning of Stage~2, the learned parameters \(\vartheta\) initialized the task predictor applied to refined concept probabilities. The task predictor and relational refinement module were subsequently optimized using the complete Stage~2 objective:
\begin{equation}
\mathcal L_{\mathrm{stage2}}^{\mathrm{complete}}
=
\mathcal L_{\mathrm{stage2}}
+\lambda_{\mathrm{logic}}\mathcal L_{\mathrm{logic}},
\label{eq:supp_complete_stage2}
\end{equation}
where we set \(\lambda_{\mathrm{logic}}=0.1\).

\clearpage

\section{Implementation Details}
\label{app:implementation_details}

This section provides the implementation and evaluation settings needed to
reproduce the experiments. The three datasets were selected to cover
complementary experimental settings. WBC provided a compact concept vocabulary
for evaluating relational refinement on medical images with interpretable
morphological attributes. CUB provided a larger setting with 112 concepts and
200 fine-grained classes, allowing us to evaluate the method with a
substantially larger concept and task space. Synthetic provided known concept
relations and a controlled generation process, allowing relational recovery to
be examined when the underlying co-occurrence, implication, and exclusion
structures were explicitly defined.

For WBC and CUB, the encoder \(f_\theta\) was a ResNet-34, and images were
resized to \(224\times224\). For Synthetic, \(f_\theta\) was an MLP with hidden
dimension 128.

ReCBM was trained for 150 epochs. Stage~1 lasted 70 epochs and jointly trained
the encoder \(f_\theta\), evidential concept predictor \(g_\psi\), and task
predictor \(h_\vartheta\). At the beginning of Stage~2, the task predictor was
initialized from Stage~1. The encoder and evidential
concept predictor were then frozen. During the remaining 80 epochs, the
relation module and task predictor were optimized.

\subsection{Training and Hyperparameter Configuration}
\label{app:hyperparameter_search}

This subsection specifies the optimization schedule and the final settings used for each dataset. Hyperparameters were selected on the validation split by task accuracy, with concept accuracy used to break ties. Refinement configurations were evaluated under the corrupted validation protocol described in Section~\ref{app:corruption_protocol}. Table~\ref{tab:supp_hyperparameters} reports the candidate values and final configuration for each dataset. In Stage~1, the encoder and concept predictor used the dataset-specific initial learning rate in Table~\ref{tab:supp_hyperparameters}, and the task predictor used the same rate. In Stage~2, the relation module and task predictor started at half this rate. Stage~1 and Stage~2 used independent cosine annealing schedules with horizons of 70 and 80 epochs, respectively, and a minimum learning rate of zero.

\begin{table*}[htbp]
    \centering
    \footnotesize
    \setlength{\tabcolsep}{3pt}
    \begin{tabular}{@{}p{0.35\textwidth}C{0.15\textwidth}C{0.15\textwidth}C{0.15\textwidth}C{0.15\textwidth}@{}}
        \toprule
        Hyperparameter & Candidate values & WBC & CUB & Synthetic \\
        \midrule
        Initial learning rate
            & $\{5{\times}10^{-4},10^{-3}\}$
            & $5{\times}10^{-4}$ & $5{\times}10^{-4}$ & $10^{-3}$ \\
        Batch size
            & $\{64,192,256\}$
            & 64 & 64 & 256 \\
        Refinement iterations
            & $\{5,10,15\}$
            & 5 & 10 & 5 \\
        Stage~1 epochs
            & $\{50,70,100\}$
            & 70 & 70 & 70 \\
        Total epochs
            & $\{100,150\}$
            & 150 & 150 & 150 \\
        Stage~2 concept-state augmentation ratio
            & $\{0,0.1\}$
            & 0.1 & 0.1 & 0.1 \\
        Natural refined-concept BCE weight
            & $\{5,10,20\}$
            & 20 & 10 & 10 \\
        Augmented refined-concept BCE weight
            & $\{1,5,10\}$
            & 5 & 10 & 10 \\
        Endpoint direction-loss weight
            & $\{1,10\}$
            & 10 & 10 & 10 \\
        Message direction-loss weight
            & $\{1,10\}$
            & 10 & 10 & 10 \\
        Uncertainty-error loss weight
            & $\{1,10\}$
            & 10 & 10 & 10 \\
        Logic-conflict weight $\lambda_{\mathrm{logic}}$
            & $\{0.1,1\}$
            & 0.1 & 0.1 & 0.1 \\
        \midrule
        Optimizer
            & AdamW
            & AdamW & AdamW & AdamW \\
        Weight decay
            & $5{\times}10^{-6}$
            & $5{\times}10^{-6}$ & $5{\times}10^{-6}$ & $5{\times}10^{-6}$ \\
        Learning-rate scheduler
            & cosine annealing
            & cosine annealing
            & cosine annealing
            & cosine annealing \\
        KL coefficient / annealing horizon
            & $1/10$ epochs
            & $1/10$ epochs & $1/10$ epochs & $1/10$ epochs \\
        Stage~1 / Stage~2 task-loss weights
            & $1/1$
            & $1/1$ & $1/1$ & $1/1$ \\
        Concept-loss weight
            & 1
            & 1 & 1 & 1 \\
        Minimum support count $\tau$
            & 5
            & 5 & 5 & 5 \\
        Wilson interval normal quantile
            & 1.96
            & 1.96 & 1.96 & 1.96 \\
        \bottomrule
    \end{tabular}
    \caption{Candidate hyperparameter values and final configurations selected
    on the validation split for each dataset.}
    \label{tab:supp_hyperparameters}
\end{table*}

\subsection{Comparison Method Configuration}
\label{app:comparison_configuration}

The comparison methods used the same dataset splits, input preprocessing,
backbones, and batch sizes as ReCBM. Image models used ResNet-34, and models
for Synthetic used the MLP encoder with hidden dimension 128. Table~
\ref{tab:supp_comparison_hyperparameters} reports the final training
configuration of each comparison method.

\begin{table*}[htbp]
    \centering
    \footnotesize
    \setlength{\tabcolsep}{3pt}
    \begin{tabular}{@{}L{0.14\textwidth}C{0.10\textwidth}
        C{0.10\textwidth}C{0.15\textwidth}L{0.43\textwidth}@{}}
        \toprule
        Method & Optimizer & Learning rate & Training epochs
        & Additional configuration \\
        \midrule
        Independent CBM
        & SGD
        & $0.01/0.01$
        & $100/100$
        & Concept and task stages used weight decay
          $4{\times}10^{-5}$. The StepLR step size was 1000 with a decay
          factor of 0.1. \\
        Joint CBM
        & SGD
        & $0.01$
        & 100
        & Weight decay was $4{\times}10^{-5}$. The concept loss weight was
          0.01, and the normalized joint loss was used. The StepLR step size
          was 1000 with a decay factor of 0.1. \\
        ProbCBM
        & AdamP
        & $10^{-3}$
        & $50/20$
        & The concept and task stages lasted 50 and 20 epochs. The first five
          epochs used encoder warmup. Newly introduced parameter groups used
          a learning rate multiplier of 10. The model used 50 Monte Carlo
          samples, concept hidden dimension 16, task hidden dimension 128,
          intervention probability 0.5, and variational weight
          $5{\times}10^{-5}$. Weight decay was zero, and each stage used
          cosine annealing. \\
        SCBM
        & Adam
        & $10^{-4}$
        & $100/300/300$
        & The epoch counts correspond to WBC, CUB, and Synthetic,
          respectively. SCBM used amortized covariance, 100 Monte Carlo
          samples, hard concept learning with the straight through estimator,
          and an $\ell_1$ precision regularizer with weight 1. Weight decay
          was zero. StepLR used a step size of 150 and a decay factor of 0.5. \\
        GraphCBM
        & Adam
        & $10^{-3}$
        & 150
        & Weight decay was $4{\times}10^{-5}$. The model used three graph
          layers, concept loss weight 1, graph regularization weight 0.1, and
          gradient clipping at 0.5. Validation was performed every five epochs
          with patience 20. The StepLR step size was 1000 with a decay factor
          of 0.1. \\
        \bottomrule
    \end{tabular}
    \caption{Final training configurations of the comparison methods. For
    Independent CBM and ProbCBM, values separated by a slash correspond to
    the concept and task training stages. For SCBM, the three epoch counts
    correspond to WBC, CUB, and Synthetic.}
    \label{tab:supp_comparison_hyperparameters}
\end{table*}

\subsection{Corruption and Evaluation Protocol}
\label{app:corruption_protocol}

For the corruption robustness experiments, the same entry-level corruption mask was used for all methods or ablation variants compared in a run. A missing entry was replaced by \((p,u)=(0.5,1)\). To flip an entry, we first converted its raw probability into a binary prediction using a threshold of 0.5 and then reversed the prediction, changing 0 to 1 and 1 to 0. Its uncertainty was subsequently set to \(u=1\). Entries outside the corruption mask retained their original probabilities and uncertainties. Thus, these experiments evaluated recovery when the locations of the corrupted concepts were explicitly identified as unreliable.

Concept accuracy was computed by thresholding concept probabilities at 0.5, whereas task accuracy was computed from the final class prediction. Standard performance results used five seeds (42-46). Ablations used three paired seeds (0-2), shared corruption masks, and a 50\% corruption ratio.

\subsection{Random Seeds and Deterministic Execution}
\label{app:random_seeds}

ReCBM training used the seed specified in each configuration through the
PyTorch Lightning seeding utility. The comparison method and evaluation
utilities explicitly seeded Python, NumPy, PyTorch, and all available CUDA
devices. For evaluation, cuDNN benchmarking was disabled, deterministic cuDNN
execution was enabled, and deterministic PyTorch operations were requested.

\subsection{Computing Environment}
\label{app:computing_environment}

Experiments were conducted on Ubuntu Linux using an AMD EPYC 7742 CPU,
256~GiB of system memory, and one NVIDIA GeForce RTX 3090 GPU with 24~GiB
of memory. The software environment consisted of Python 3.9.20, PyTorch 2.5.0, torchvision 0.20.0, PyTorch Lightning 2.6.0, CUDA 12.4, cuDNN 9.1, NumPy 2.0.1, and scikit-learn 1.5.2.  The code archive contains a version-pinned requirements file.

\clearpage
\section{ReCBM Training Pseudocode}
\label{app:pseudocode}

This section presents the ReCBM training procedure with two stages.
In Algorithm~\ref{alg:recbm_training},
\(\mathcal D=\{(x_n,y_n,\mathbf c_n)\}_{n=1}^{N}\) is the training set,
\(\mathcal A=\{\mathbf A^{\mathrm{co}},\mathbf A^{\mathrm{imp}},
\mathbf A^{\mathrm{exc}}\}\), \(E_1\) and \(E\) are the numbers of Stage~1
and total epochs, and \(T\) is the number of refinement iterations. The
encoder, concept predictor, relation module, and task predictor are denoted
by \(f_\theta\), \(g_\psi\), \(r_\phi\), and \(h_\vartheta\), respectively.
The concept predictor outputs raw probabilities
\(\mathbf p^{\mathrm{raw}}\) and uncertainties \(\mathbf u\), while
\(\mathbf p^{(t)}\) denotes the concept state at refinement iteration \(t\).

\begin{algorithm*}[htbp]
\caption{Training of ReCBM in two stages}
\label{alg:recbm_training}
\begin{algorithmic}[1]
\REQUIRE Data $\mathcal D$; relation matrices $\mathcal A$; Stage-1 epochs
$E_1$; total epochs $E$; refinement steps $T$
\STATE Initialize encoder $f_\theta$, concept predictor $g_\psi$, task
predictor $h_\vartheta$, and relation module $r_\phi$
\FOR{$e=1,\ldots,E_1$}
    \FOR{each minibatch $(x,y,\mathbf c)$}
        \STATE $(\mathbf p^{\mathrm{raw}},\mathbf u)\leftarrow
        g_\psi(f_\theta(x))$
        \STATE Update $(\theta,\psi,\vartheta)$ with the Stage-1 loss
    \ENDFOR
\ENDFOR
\STATE Retain the learned $\vartheta$ as the Stage~2 initialization and
freeze $f_\theta$ and $g_\psi$
\FOR{$e=E_1+1,\ldots,E$}
    \FOR{each minibatch $(x,y,\mathbf c)$}
        \STATE $(\mathbf p^{\mathrm{raw}},\mathbf u)\leftarrow
        g_\psi(f_\theta(x))$;
        $(\mathbf p^{(0)},\mathbf u^{(0)})\leftarrow
        \operatorname{Augment}(\mathbf p^{\mathrm{raw}},\mathbf u)$
        \FOR{$t=0,\ldots,T-1$}
            \STATE $\mathbf p^{(t+1)}\leftarrow
            r_\phi(\mathbf p^{(t)},\mathbf u^{(0)},\mathcal A)$
        \ENDFOR
        \STATE $\mathbf p^{\mathrm{ref}}\leftarrow
        \operatorname{Anchor}(\mathbf p^{(0)},\mathbf p^{(T)},
        \mathbf u^{(0)},\mathcal A)$
        \STATE Update $(\phi,\vartheta)$ with
        Eq.~\eqref{eq:supp_complete_stage2}
    \ENDFOR
\ENDFOR
\RETURN Trained parameters \(\theta,\psi,\phi,\vartheta\)
\end{algorithmic}
\end{algorithm*}
\FloatBarrier

\clearpage

\section{Complete Ablation Results}
\label{app:complete_ablation}

This section evaluates the contribution of each relation type and the
uncertainty gate across all three datasets.
Table~\ref{tab:supp_ablation_50} extends the WBC table in the main paper to all three datasets while retaining both corruption types. It reports results using the same three paired seeds and a corruption ratio of 50\%.

\begin{table*}[htbp]
    \centering
    \small
    \setlength{\tabcolsep}{2.2pt}
    \begin{tabular}{@{}C{0.08\textwidth}C{0.15\textwidth}
        *{4}{C{0.045\textwidth}}C{0.24\textwidth}C{0.24\textwidth}@{}}
        \toprule
        Dataset & Variant & Co. & Im. & Ex. & Gate & Flip & Missing \\
        \midrule
        \multirow{9}{*}{WBC}
        & Full ReCBM & \checkmark&\checkmark&\checkmark&\checkmark & \textbf{80.10 $\pm$ 3.14 / 76.00 $\pm$ 2.62} & \textbf{90.89 $\pm$ 0.21 / 95.31 $\pm$ 0.49} \\
        & w/o Co. &$\times$&\checkmark&\checkmark&\checkmark &74.53 $\pm$ 1.70 / 59.75 $\pm$ 1.42&90.61 $\pm$ 0.07 / 94.87 $\pm$ 0.26\\
        & w/o Im. &\checkmark&$\times$&\checkmark&\checkmark &72.89 $\pm$ 0.21 / 71.81 $\pm$ 0.50&88.70 $\pm$ 0.15 / 94.84 $\pm$ 0.69\\
        & w/o Ex. &\checkmark&\checkmark&$\times$&\checkmark &78.09 $\pm$ 5.03 / 65.87 $\pm$ 7.34&89.23 $\pm$ 0.36 / 94.85 $\pm$ 0.53\\
        & Co. only &\checkmark&$\times$&$\times$&\checkmark &64.53 $\pm$ 1.46 / 55.02 $\pm$ 1.12&86.42 $\pm$ 0.14 / 93.97 $\pm$ 0.40\\
        & Im. only &$\times$&\checkmark&$\times$&\checkmark &59.24 $\pm$ 0.97 / 30.54 $\pm$ 1.81&85.69 $\pm$ 0.05 / 93.99 $\pm$ 0.47\\
        & Ex. only &$\times$&$\times$&\checkmark&\checkmark &66.02 $\pm$ 0.25 / 50.31 $\pm$ 1.05&81.20 $\pm$ 0.07 / 94.30 $\pm$ 0.41\\
        & w/o Relations &$\times$&$\times$&$\times$&\checkmark &49.94 $\pm$ 0.11 / 13.92 $\pm$ 0.63&64.44 $\pm$ 0.14 / 91.84 $\pm$ 0.86\\
        & w/o Gating &\checkmark&\checkmark&\checkmark&$\times$ &74.83 $\pm$ 0.06 / 14.71 $\pm$ 0.99&85.77 $\pm$ 0.12 / 90.59 $\pm$ 1.85\\
        \midrule
        \multirow{9}{*}{CUB}
        & Full ReCBM &\checkmark&\checkmark&\checkmark&\checkmark &\textbf{87.65 $\pm$ 0.07 / 40.09 $\pm$ 0.27}&91.51 $\pm$ 0.14 / \textbf{67.31 $\pm$ 0.28}\\
        & w/o Co. &$\times$&\checkmark&\checkmark&\checkmark &84.45 $\pm$ 0.49 / 28.74 $\pm$ 2.28&90.24 $\pm$ 0.07 / 67.24 $\pm$ 0.53\\
        & w/o Im. &\checkmark&$\times$&\checkmark&\checkmark &85.81 $\pm$ 0.25 / 38.58 $\pm$ 2.46&90.00 $\pm$ 0.48 / 65.69 $\pm$ 0.84\\
        & w/o Ex. &\checkmark&\checkmark&$\times$&\checkmark &86.25 $\pm$ 0.24 / 37.13 $\pm$ 3.20&\textbf{91.54 $\pm$ 0.18} / 67.13 $\pm$ 1.13\\
        & Co. only &\checkmark&$\times$&$\times$&\checkmark &83.00 $\pm$ 0.23 / 33.70 $\pm$ 0.64&89.92 $\pm$ 0.06 / 66.22 $\pm$ 0.68\\
        & Im. only &$\times$&\checkmark&$\times$&\checkmark &78.72 $\pm$ 1.39 / 17.32 $\pm$ 4.43&87.84 $\pm$ 0.38 / 66.35 $\pm$ 0.54\\
        & Ex. only &$\times$&$\times$&\checkmark&\checkmark &79.05 $\pm$ 0.96 / 10.73 $\pm$ 1.16&87.78 $\pm$ 0.06 / 66.80 $\pm$ 1.40\\
        & w/o Relations &$\times$&$\times$&$\times$&\checkmark &50.03 $\pm$ 0.02 / 0.12 $\pm$ 0.06&58.02 $\pm$ 0.29 / 66.24 $\pm$ 1.47\\
        & w/o Gating &\checkmark&\checkmark&\checkmark&$\times$ &74.52 $\pm$ 0.07 / 0.32 $\pm$ 0.05&86.18 $\pm$ 0.06 / 16.57 $\pm$ 0.99\\
        \midrule
        \multirow{9}{*}{Synthetic}
        & Full ReCBM &\checkmark&\checkmark&\checkmark&\checkmark &\textbf{84.23 $\pm$ 1.06 / 77.34 $\pm$ 2.07}&\textbf{91.27 $\pm$ 0.28 / 89.97 $\pm$ 0.15}\\
        & w/o Co. &$\times$&\checkmark&\checkmark&\checkmark &68.85 $\pm$ 8.97 / 58.00 $\pm$ 9.74&85.82 $\pm$ 0.13 / 89.19 $\pm$ 0.52\\
        & w/o Im. &\checkmark&$\times$&\checkmark&\checkmark &82.55 $\pm$ 1.35 / 76.49 $\pm$ 2.44&90.69 $\pm$ 0.23 / 89.76 $\pm$ 0.16\\
        & w/o Ex. &\checkmark&\checkmark&$\times$&\checkmark &79.23 $\pm$ 0.56 / 65.78 $\pm$ 2.45&87.75 $\pm$ 0.86 / 89.53 $\pm$ 0.28\\
        & Co. only &\checkmark&$\times$&$\times$&\checkmark &77.21 $\pm$ 0.50 / 64.80 $\pm$ 2.94&87.39 $\pm$ 0.82 / 89.07 $\pm$ 0.73\\
        & Im. only &$\times$&\checkmark&$\times$&\checkmark &55.67 $\pm$ 4.63 / 35.13 $\pm$ 4.32&81.29 $\pm$ 0.52 / 89.09 $\pm$ 0.41\\
        & Ex. only &$\times$&$\times$&\checkmark&\checkmark &61.16 $\pm$ 8.61 / 49.53 $\pm$ 9.31&78.42 $\pm$ 0.13 / 88.93 $\pm$ 0.37\\
        & w/o Relations &$\times$&$\times$&$\times$&\checkmark &49.99 $\pm$ 0.35 / 24.74 $\pm$ 2.35&70.72 $\pm$ 1.55 / 88.97 $\pm$ 0.15\\
        & w/o Gating &\checkmark&\checkmark&\checkmark&$\times$ &57.83 $\pm$ 1.37 / 24.64 $\pm$ 2.90&89.28 $\pm$ 0.10 / 85.16 $\pm$ 0.69\\
        \bottomrule
    \end{tabular}
    \caption{Complete component ablation under 50\% concept corruption.
    Entries are concept/task accuracy (\%), reported as mean $\pm$ standard
    deviation over three paired seeds.  Bold denotes the best mean for each
    dataset and corruption type.}
    \label{tab:supp_ablation_50}
\end{table*}

\clearpage
\section{Qualitative Refinement Analysis}
\label{subsec:qualitative_analysis}

\begin{figure*}[htbp]
    \centering
    \begin{subfigure}[t]{0.42\textwidth}
        \centering
        \includegraphics[width=\linewidth]
        {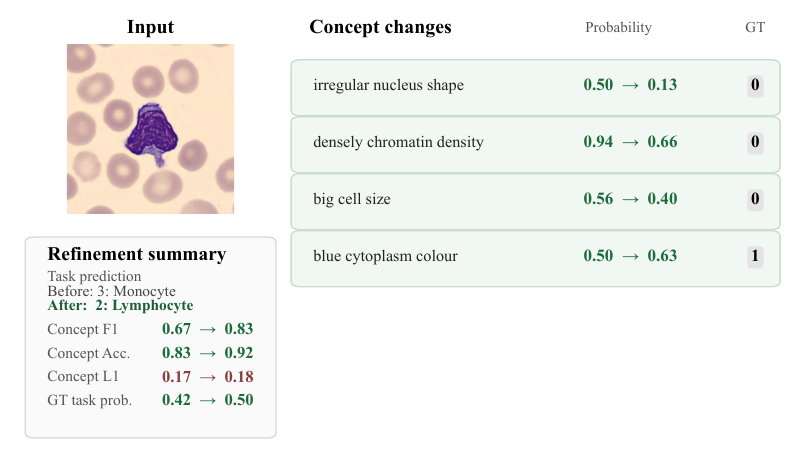}\\[3pt]
        \includegraphics[width=\linewidth]
        {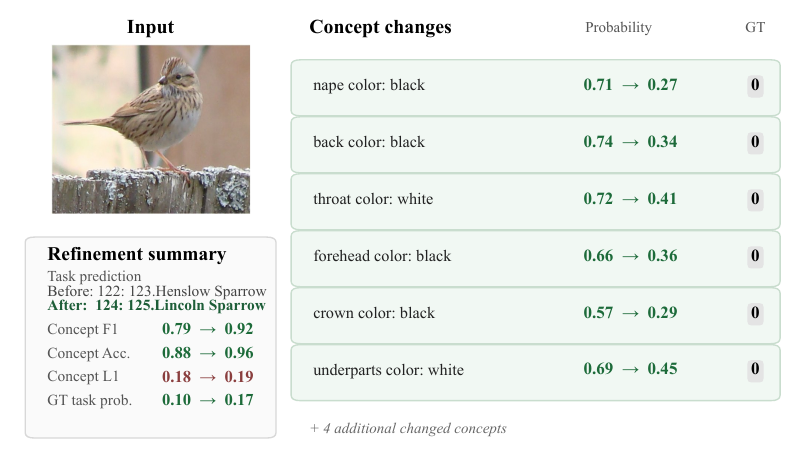}\\[3pt]
        \includegraphics[width=\linewidth]
        {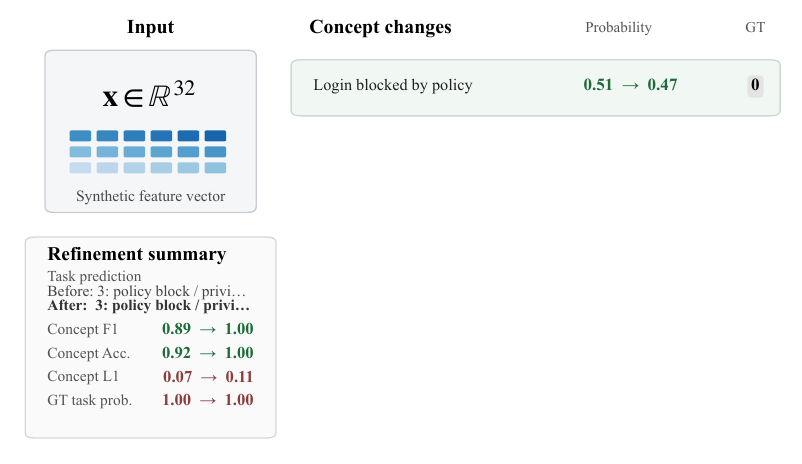}
        \caption{Refinement of uncorrupted concept predictions.}
        \label{fig:supp_raw_cases}
    \end{subfigure}\hfill
    \begin{subfigure}[t]{0.42\textwidth}
        \centering
        \includegraphics[width=\linewidth]
        {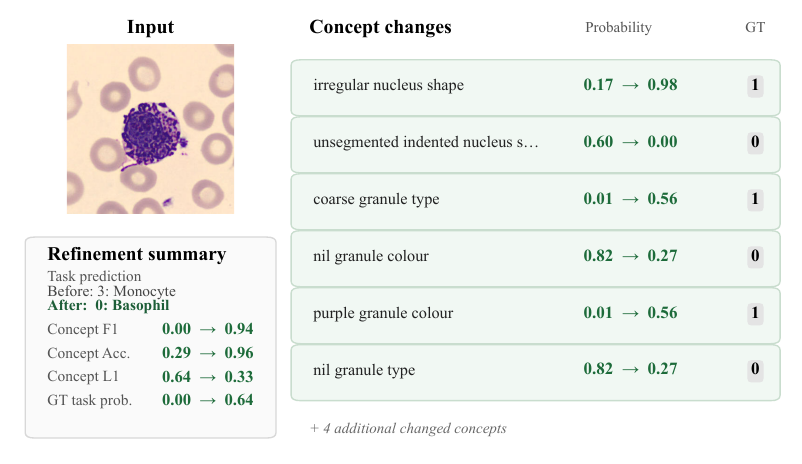}\\[3pt]
        \includegraphics[width=\linewidth]
        {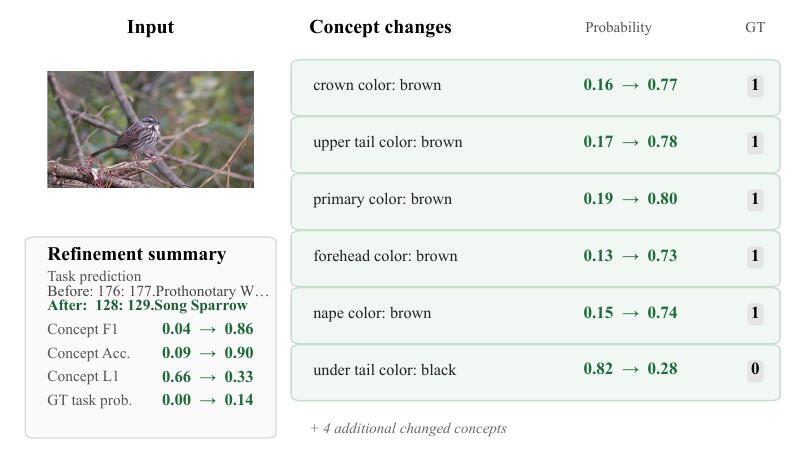}\\[3pt]
        \includegraphics[width=\linewidth]
        {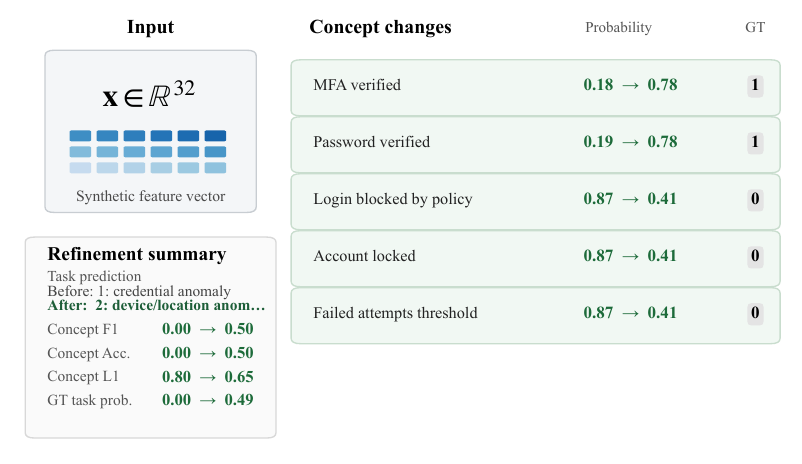}
        \caption{Refinement of corrupted concept predictions after marking
        corrupted entries as uncertain.}
        \label{fig:supp_uncertainty_cases}
    \end{subfigure}
    \caption{Representative qualitative refinement examples on WBC, CUB, and
    Synthetic. Left: concept predictions were refined without applying
    corruption. Refinement improved concept F1 from 0.667 to 0.833 on WBC,
    from 0.787 to 0.915 on CUB, and from 0.889 to 1.000 on Synthetic. It
    corrected the WBC and CUB task predictions while leaving the already
    correct Synthetic prediction unchanged. Right: concept predictions were
    first corrupted, after which the corrupted entries were marked as
    unreliable by setting only their uncertainty to one, and their correct concept
    values were not provided. Refinement improved concept F1 from 0 to 0.941
    on WBC, from 0.038 to 0.857 on CUB, and from 0 to 0.500 on Synthetic, and
    corrected the task predictions in all three cases.}
    \label{fig:supp_qualitative_combined}
\end{figure*}

This section illustrates how relational refinement changed concept and task
predictions in representative examples.
Figures~\ref{fig:supp_raw_cases} and \ref{fig:supp_uncertainty_cases} present representative refinement cases. Each dataset is shown in a separate row to preserve readability. Without concept corruption, refinement increased concept F1 from 0.667 to 0.833 on WBC, from 0.787 to 0.915 on CUB, and from 0.889 to 1.000 on Synthetic. It corrected the WBC and CUB task predictions while leaving the already correct Synthetic prediction unchanged.

For the uncertainty-only cases, only the uncertainty assigned to unreliable concepts was changed, and no correct concept values were supplied. Relational refinement increased concept F1 from 0 to 0.941 on WBC, from 0.038 to 0.857 on CUB, and from 0 to 0.500 on Synthetic, while correcting the task prediction in all three cases.

\FloatBarrier

\clearpage
\section{Uncertainty-Only Intervention}
\label{app:uncertainty_recovery}

This section examines whether changes in uncertainty alone could improve concept recovery while leaving the concept probabilities unchanged. We first divided raw concept predictions into two candidate groups. The first group contained correct predictions with uncertainty no greater than 0.2, and the second contained incorrect predictions with uncertainty no less than 0.8. Within each group, 30\% of the eligible predictions were sampled and flipped by reversing their thresholded binary values. This produced incorrect predictions with low uncertainty and correct predictions with high uncertainty.

For each intervention ratio, we sampled the corresponding proportion of the first group and set their uncertainty to \(u=1\). We sampled the same proportion of the second group and set their uncertainty to \(u=0\). The corrupted concept probabilities remained fixed throughout the intervention. Any improvement therefore resulted from changing the reliability information provided to the refinement module rather than supplying corrected concept values.

The uncertainty intervention figure in the main paper reports the results for WBC, CUB, and Synthetic. It shows that concept and task performance improved as the intervention
adjusted the uncertainty of a larger proportion of the sampled concepts,
assigning high uncertainty to incorrect predictions. Figure~\ref{fig:supp_uncertainty_cases} provides individual examples of how these changes affected the refined concept predictions.

\section{Counterfactual Edits under Different Uncertainty Levels}
\label{app:uncertainty_counterfactual}

This section tests whether uncertainty regulated the effect of identical
counterfactual concept edits on task predictions.

We considered test examples that were correctly classified by the complete
model. For each example, the alternative class was the class other than the original
prediction that had the highest original probability. For binary
classification, this was the class opposite to the original prediction.

Candidate concepts were scored using both their relational strength and their
effect on the task logits. For each concept, we first reversed its thresholded
raw prediction and set its uncertainty to zero. We then measured the sum of
the decrease in the original class logit and the increase in the alternative
class logit, with negative values clipped to zero. The relational score was
the strongest learned relation involving that concept, considering
co-occurrence, exclusion, and both directions of implication. After separately
normalizing the logit shift and relational scores to \([0,1]\), we multiplied
them and selected the \(K\in\{1,3,5\}\) concepts with the highest resulting
scores.

For each example and edit budget, the selected concept indices and their
edited values were identical in the two uncertainty settings. Each selected
raw prediction was converted to a binary value using a threshold of 0.5 and
then reversed. The uncertainty of every edited concept was set either to
\(u=0\), indicating a reliable edit, or to \(u=1\), indicating an unreliable
edit. Target success was the fraction of examples whose final prediction
changed to the selected alternative class.
Table~\ref{tab:supp_counterfactual} reports the results.
\begin{table}[htbp]
    \centering
    \small
    \setlength{\tabcolsep}{4pt}
    \begin{tabular}{@{}C{0.20\textwidth}*{4}{C{0.14\textwidth}}@{}}
        \toprule
        Dataset & $K$ & Low $u$ & High $u$ & Difference \\
        \midrule
        \multirow{3}{*}{WBC}
        &1&7.02&1.82&5.20\\
        &3&97.01&20.92&76.09\\
        &5&99.97&84.70&15.27\\
        \midrule
        \multirow{3}{*}{CUB}
        &1&18.44&5.27&13.17\\
        &3&48.57&16.09&32.48\\
        &5&80.12&29.72&50.40\\
        \midrule
        \multirow{3}{*}{Synthetic}
        &1&19.49&3.65&15.84\\
        &3&99.01&30.85&68.16\\
        &5&98.81&91.06&7.75\\
        \bottomrule
    \end{tabular}
    \caption{Target class success (\%) of counterfactual concept edits under low and high uncertainty. \(K\) denotes the number of edited concepts, and Difference denotes the low uncertainty result minus the high uncertainty result.}
    \label{tab:supp_counterfactual}
\end{table}

The desired behavior differed between the two settings. Under low uncertainty, a high target success rate indicated that reliable concept edits effectively controlled the task prediction. Under high uncertainty, a low target success rate indicated that the model resisted the same edits after they were marked as unreliable. Target success generally increased with \(K\), while remaining lower under high uncertainty. The gap between the two settings therefore measured how strongly uncertainty regulated the influence of identical concept edits. This behavior also provided a degree of robustness to potentially incorrect human interventions: when an edit was marked as uncertain, the model reduced its influence and relied more on the remaining reliable concepts and learned relations.

\clearpage
\section{Sufficient Concept-Set Extraction}
\label{app:concept_set_extraction}

This section identifies compact concept sets that retained the predictive
performance obtained from the complete concept representation. A global set
provides a compact concept interface for the overall task, whereas a set
selected for an individual class identifies the concepts sufficient for
predictions associated with that class. If a small subset achieves
performance comparable to that obtained using all concepts, users need to
inspect or provide feedback for fewer concepts, reducing the burden of human
interaction. All set selection was performed on the validation split.

During set selection, the selected dimensions retained their predicted raw
probabilities and were assigned \(u=0\), while all unselected dimensions were
replaced by the unknown state \((p,u)=(0.5,1)\). ReCBM then reconstructed the
complete refined concept vector before applying the task predictor. The same
input protocol was used to evaluate the global set on the test split. For the
final evaluation of each class-specific set, however, the selected dimensions
were supplied with their ground truth binary values and assigned \(u=0\).
This setting more closely represented practical human intervention, in which
users provide definite concept values rather than continuous prediction
probabilities. The unselected dimensions remained in the unknown state.

To obtain the global set, we ranked concept dimensions by the product of
their mean absolute task logit gradient and mean anchor gate. The gradient
was computed for the logit corresponding to the ground truth task label on
the validation split. We searched the ranked prefixes for the smallest
prefix satisfying both retention criteria. Starting from this prefix, we
considered concepts in increasing order of their ranking scores and removed
a concept whenever both criteria remained satisfied.

For a candidate set \(S\), concept retention and task retention were defined
as
\begin{equation}
R_{\mathrm{concept}}(S)
=
\frac{\operatorname{F1}_{\mathrm{concept}}(S)}
{\operatorname{F1}_{\mathrm{concept}}(\mathrm{full})},
\qquad
R_{\mathrm{task}}(S)
=
\frac{\operatorname{Acc}_{\mathrm{task}}(S)}
{\operatorname{Acc}_{\mathrm{task}}(\mathrm{full})}.
\end{equation}
A set satisfied a retention target \(\gamma\) only when
\(R_{\mathrm{concept}}(S)\geq\gamma\) and
\(R_{\mathrm{task}}(S)\geq\gamma\).

For each individual class, selection was performed using only validation
examples from that class. Starting from an empty set, we evaluated every
unselected concept as the next candidate and added the concept that produced
the greatest minimum of concept F1 retention and task accuracy retention.
This process continued until both retention criteria reached
\(\gamma=1\). We then removed selected concepts one at a time whenever both
criteria remained satisfied. The resulting concept sets were fixed before
their final evaluation on the test split.

\begin{table}[htbp]
    \centering
    \small
    \setlength{\tabcolsep}{4pt}
    \begin{tabular}{@{}C{0.20\textwidth}*{4}{C{0.14\textwidth}}@{}}
        \toprule
        Dataset & 90\% & 95\% & 98\% & 100\% \\
        \midrule
        WBC &16/24&16/24&17/24&18/24\\
        CUB &64/112&88/112&94/112&99/112\\
        Synthetic &8/12&11/12&11/12&12/12\\
        \bottomrule
    \end{tabular}
    \caption{Sizes of the global concept sets selected on the validation split at additional retention targets. Selection required both concept F1 and task accuracy retention to meet the stated target.}
    \label{tab:supp_retention_thresholds}
\end{table}

Table~\ref{tab:supp_retention_thresholds} shows how the number of globally selected concepts changed with the required performance retention level across the three datasets. Tables~\ref{tab:supp_class_sets_wbc}, \ref{tab:supp_class_sets_synthetic}, and \ref{tab:supp_class_sets_cub112} report the concept sets selected separately for each class. Selection was performed on the validation split, and only sets that satisfied the retention criteria were included. Valid sets were obtained for all five WBC classes, all four Synthetic classes, and 166 of the 200 CUB classes. These sets were subsequently evaluated on the test split. The results show that ReCBM identified compact concept sets that satisfied both retention criteria on the validation split and largely preserved performance on the test split. These sets required users to inspect or intervene on only a subset of the available concepts. The global sets provided a compact concept interface for the overall task, whereas the sets selected for individual classes provided more focused interfaces for inspection and intervention within each class.

\begin{table*}[htbp]
    \centering
    \footnotesize
    \setlength{\tabcolsep}{3pt}
    \begin{tabular}{@{}C{0.17\textwidth}C{0.07\textwidth}C{0.68\textwidth}@{}}
        \toprule
        Class & Size & Selected concepts \\
        \midrule
        Basophil & 6 & purple granule colour; big cell size; irregular nucleus shape; segmented multilobed nucleus shape; round cell shape; densely chromatin density \\
        Eosinophil & 5 & red granule colour; purple granule colour; segmented multilobed nucleus shape; round cell shape; small granule type \\
        Lymphocyte & 11 & high nuclear cytoplasmic ratio; purple blue cytoplasm colour; big cell size; round cell shape; irregular nucleus shape; clear cytoplasm texture; round granule type; red granule colour; light blue cytoplasm colour; granularity; densely chromatin density \\
        Monocyte & 8 & unsegmented indented nucleus shape; high nuclear cytoplasmic ratio; cytoplasm vacuole; nil granule type; round cell shape; purple blue cytoplasm colour; unsegmented round nucleus shape; big cell size \\
        Neutrophil & 3 & unsegmented band nucleus shape; small granule type; clear cytoplasm texture \\
        \bottomrule
    \end{tabular}
    \caption{Class-specific sufficient concept sets for WBC.}
    \label{tab:supp_class_sets_wbc}
\end{table*}

\begin{table*}[htbp]
    \centering
    \footnotesize
    \setlength{\tabcolsep}{3pt}
    \begin{tabular}{@{}C{0.17\textwidth}C{0.07\textwidth}C{0.68\textwidth}@{}}
        \toprule
        Class & Size & Selected concepts \\
        \midrule
        Normal access & 11 & Login blocked by policy; Usual location; Password verified; Administrator account; New device; Account locked; MFA verified; Administrator operation; Login succeeds; Registered device; Anomalous location \\ Credential anomaly & 8 & Failed attempts reach threshold; Anomalous location; Administrator operation; Registered device; Login blocked by policy; Password verified; New device; Account locked \\ Device or location anomaly & 12 & Anomalous location; Login blocked by policy; Usual location; Administrator account; Account locked; Password verified; Login succeeds; MFA verified; Administrator operation; Registered device; Failed attempts reach threshold; New device \\ Policy block or privilege risk & 11 & Administrator operation; Password verified; Registered device; Anomalous location; Account locked; Login blocked by policy; Login succeeds; Administrator account; Usual location; New device; MFA verified \\
        \bottomrule
    \end{tabular}
    \caption{Class-specific sufficient concept sets for Synthetic.}
    \label{tab:supp_class_sets_synthetic}
\end{table*}

\FloatBarrier
{\footnotesize
\setlength{\tabcolsep}{3pt}
\begin{longtable}{@{}C{0.17\textwidth}C{0.07\textwidth}C{0.68\textwidth}@{}}
    \caption{Class-specific sufficient concept sets for CUB.}
    \label{tab:supp_class_sets_cub112}\\
    \toprule
    Class & Size & Selected concepts \\
    \midrule
    \endfirsthead
    \multicolumn{3}{c}{\tablename\ \thetable\ (continued)}\\
    \toprule
    Class & Size & Selected concepts \\
    \midrule
    \endhead
    \midrule
    \multicolumn{3}{c}{Continued on the next page}\\
    \endfoot
    \bottomrule
    \endlastfoot
Black footed Albatross & 10 & has\_bill\_shape: all-purpose; has\_wing\_color: buff; has\_wing\_color: white; has\_upperparts\_color: black; has\_underparts\_color: white; has\_back\_color: white; has\_breast\_pattern: solid; has\_upper\_tail\_color: grey; has\_bill\_shape: hooked\_seabird; has\_head\_pattern: plain \\
        Laysan Albatross & 11 & has\_bill\_shape: hooked\_seabird; has\_nape\_color: grey; has\_wing\_color: grey; has\_wing\_color: black; has\_wing\_color: brown; has\_wing\_color: yellow; has\_upperparts\_color: brown; has\_upperparts\_color: yellow; has\_underparts\_color: brown; has\_underparts\_color: grey; has\_size: small\_(5\_-\_9\_in) \\
        Sooty Albatross & 8 & has\_belly\_pattern: solid; has\_throat\_color: grey; has\_bill\_length: shorter\_than\_head; has\_upperparts\_color: buff; has\_upper\_tail\_color: brown; has\_under\_tail\_color: brown; has\_upper\_tail\_color: black; has\_bill\_shape: dagger \\
        Crested Auklet & 3 & has\_throat\_color: black; has\_bill\_length: about\_the\_same\_as\_head; has\_underparts\_color: white \\
        Least Auklet & 4 & has\_wing\_color: black; has\_eye\_color: black; has\_breast\_pattern: striped; has\_wing\_color: white \\
        Parakeet Auklet & 10 & has\_forehead\_color: black; has\_leg\_color: black; has\_eye\_color: black; has\_forehead\_color: blue; has\_underparts\_color: buff; has\_back\_color: grey; has\_primary\_color: grey; has\_underparts\_color: brown; has\_throat\_color: white; has\_wing\_pattern: solid \\
        Rhinoceros Auklet & 7 & has\_underparts\_color: white; has\_throat\_color: white; has\_belly\_color: white; has\_size: medium\_(9\_-\_16\_in); has\_wing\_pattern: spotted; has\_wing\_color: grey; has\_bill\_length: shorter\_than\_head \\
        Red winged Blackbird & 9 & has\_wing\_color: black; has\_bill\_color: buff; has\_belly\_color: white; has\_breast\_color: yellow; has\_wing\_pattern: multi-colored; has\_leg\_color: black; has\_eye\_color: black; has\_breast\_pattern: solid; has\_tail\_shape: notched\_tail \\
        Rusty Blackbird & 14 & has\_eye\_color: black; has\_upperparts\_color: white; has\_breast\_pattern: multi-colored; has\_wing\_pattern: spotted; has\_back\_pattern: multi-colored; has\_belly\_pattern: solid; has\_bill\_shape: hooked\_seabird; has\_wing\_color: buff; has\_upperparts\_color: grey; has\_breast\_color: grey; has\_underparts\_color: white; has\_wing\_shape: rounded-wings; has\_size: small\_(5\_-\_9\_in); has\_upperparts\_color: buff \\
        Bobolink & 10 & has\_leg\_color: grey; has\_crown\_color: white; has\_throat\_color: black; has\_size: small\_(5\_-\_9\_in); has\_back\_pattern: solid; has\_belly\_color: yellow; has\_primary\_color: black; has\_bill\_shape: dagger; has\_back\_pattern: multi-colored; has\_wing\_color: white \\
        Indigo Bunting & 5 & has\_forehead\_color: blue; has\_head\_pattern: plain; has\_bill\_color: black; has\_wing\_pattern: multi-colored; has\_crown\_color: blue \\
        Lazuli Bunting & 12 & has\_forehead\_color: blue; has\_bill\_shape: all-purpose; has\_crown\_color: blue; has\_breast\_pattern: striped; has\_breast\_color: grey; has\_throat\_color: grey; has\_underparts\_color: grey; has\_belly\_color: grey; has\_primary\_color: brown; has\_wing\_shape: rounded-wings; has\_wing\_pattern: multi-colored; has\_breast\_color: buff \\
        Painted Bunting & 7 & has\_forehead\_color: blue; has\_nape\_color: grey; has\_breast\_color: white; has\_forehead\_color: brown; has\_forehead\_color: black; has\_bill\_shape: all-purpose; has\_shape: perching-like \\
        Cardinal & 10 & has\_bill\_color: black; has\_upper\_tail\_color: black; has\_wing\_pattern: multi-colored; has\_forehead\_color: black; has\_breast\_color: black; has\_throat\_color: white; has\_forehead\_color: blue; has\_back\_pattern: multi-colored; has\_underparts\_color: yellow; has\_wing\_color: grey \\
        Spotted Catbird & 10 & has\_throat\_color: buff; has\_bill\_color: black; has\_bill\_shape: dagger; has\_wing\_color: black; has\_leg\_color: grey; has\_breast\_pattern: solid; has\_size: small\_(5\_-\_9\_in); has\_belly\_color: buff; has\_under\_tail\_color: grey; has\_wing\_color: yellow \\
        Gray Catbird & 5 & has\_nape\_color: grey; has\_underparts\_color: yellow; has\_bill\_shape: hooked\_seabird; has\_upperparts\_color: brown; has\_underparts\_color: white \\
        Yellow breasted Chat & 8 & has\_wing\_color: black; has\_leg\_color: grey; has\_forehead\_color: blue; has\_upper\_tail\_color: grey; has\_tail\_pattern: solid; has\_primary\_color: grey; has\_throat\_color: yellow; has\_wing\_color: yellow \\
        Eastern Towhee & 7 & has\_breast\_pattern: multi-colored; has\_bill\_shape: all-purpose; has\_underparts\_color: grey; has\_primary\_color: grey; has\_back\_color: grey; has\_throat\_color: black; has\_size: small\_(5\_-\_9\_in) \\
        Chuck will Widow & 6 & has\_underparts\_color: brown; has\_primary\_color: buff; has\_upperparts\_color: grey; has\_size: very\_small\_(3\_-\_5\_in); has\_wing\_shape: rounded-wings; has\_wing\_shape: pointed-wings \\
        Brandt Cormorant & 2 & has\_throat\_color: black; has\_bill\_shape: all-purpose \\
Red faced Cormorant & 15 & has\_belly\_color: black; has\_primary\_color: brown; has\_upper\_tail\_color: white; has\_bill\_shape: all-purpose; has\_wing\_shape: rounded-wings; has\_bill\_shape: cone; has\_wing\_color: brown; has\_breast\_pattern: solid; has\_wing\_color: grey; has\_head\_pattern: plain; has\_underparts\_color: black; has\_wing\_color: buff; has\_upperparts\_color: brown; has\_upperparts\_color: grey; has\_throat\_color: white \\
        Fish Crow & 14 & has\_tail\_pattern: solid; has\_wing\_color: grey; has\_leg\_color: black; has\_bill\_shape: cone; has\_underparts\_color: black; has\_wing\_shape: rounded-wings; has\_upperparts\_color: grey; has\_breast\_pattern: solid; has\_breast\_pattern: multi-colored; has\_back\_color: brown; has\_back\_color: grey; has\_back\_color: black; has\_wing\_color: white; has\_throat\_color: black \\
        Black billed Cuckoo & 8 & has\_eye\_color: black; has\_belly\_color: brown; has\_primary\_color: buff; has\_upperparts\_color: black; has\_head\_pattern: eyebrow; has\_breast\_color: grey; has\_primary\_color: brown; has\_back\_pattern: solid \\
        Mangrove Cuckoo & 6 & has\_throat\_color: yellow; has\_leg\_color: buff; has\_breast\_color: grey; has\_forehead\_color: grey; has\_underparts\_color: buff; has\_bill\_shape: cone \\
        Yellow billed Cuckoo & 5 & has\_eye\_color: black; has\_back\_color: black; has\_leg\_color: grey; has\_belly\_color: white; has\_crown\_color: white \\
        Purple Finch & 7 & has\_bill\_color: buff; has\_upperparts\_color: grey; has\_upperparts\_color: black; has\_size: very\_small\_(3\_-\_5\_in); has\_tail\_shape: notched\_tail; has\_underparts\_color: white; has\_nape\_color: brown \\
        Northern Flicker & 11 & has\_throat\_color: black; has\_wing\_color: black; has\_wing\_pattern: multi-colored; has\_underparts\_color: white; has\_forehead\_color: yellow; has\_underparts\_color: brown; has\_wing\_pattern: spotted; has\_leg\_color: black; has\_leg\_color: grey; has\_wing\_color: buff; has\_wing\_shape: pointed-wings \\
        Great Crested Flycatcher & 8 & has\_forehead\_color: blue; has\_underparts\_color: white; has\_wing\_pattern: spotted; has\_belly\_color: white; has\_wing\_color: grey; has\_breast\_color: grey; has\_back\_color: black; has\_throat\_color: grey \\
        Least Flycatcher & 15 & has\_eye\_color: black; has\_belly\_color: white; has\_primary\_color: black; has\_leg\_color: grey; has\_under\_tail\_color: grey; has\_underparts\_color: grey; has\_bill\_color: black; has\_nape\_color: buff; has\_wing\_pattern: striped; has\_belly\_pattern: solid; has\_bill\_shape: hooked\_seabird; has\_upper\_tail\_color: white; has\_bill\_length: about\_the\_same\_as\_head; has\_bill\_length: shorter\_than\_head; has\_breast\_color: yellow \\
        Olive sided Flycatcher & 5 & has\_eye\_color: black; has\_wing\_color: white; has\_under\_tail\_color: buff; has\_breast\_pattern: multi-colored; has\_crown\_color: yellow \\
        Scissor tailed Flycatcher & 28 & has\_bill\_shape: hooked\_seabird; has\_bill\_color: buff; has\_nape\_color: buff; has\_breast\_color: brown; has\_throat\_color: buff; has\_under\_tail\_color: buff; has\_belly\_color: brown; has\_size: very\_small\_(3\_-\_5\_in); has\_tail\_pattern: striped; has\_underparts\_color: grey; has\_belly\_color: grey; has\_forehead\_color: blue; has\_crown\_color: blue; has\_back\_pattern: multi-colored; has\_bill\_color: grey; has\_back\_pattern: striped; has\_bill\_shape: cone; has\_wing\_color: brown; has\_upperparts\_color: brown; has\_underparts\_color: buff; has\_back\_color: buff; has\_breast\_color: buff; has\_throat\_color: yellow; has\_belly\_color: buff; has\_tail\_shape: notched\_tail; has\_size: small\_(5\_-\_9\_in); has\_under\_tail\_color: grey; has\_back\_color: black \\
        Vermilion Flycatcher & 12 & has\_breast\_pattern: solid; has\_size: small\_(5\_-\_9\_in); has\_bill\_shape: hooked\_seabird; has\_primary\_color: white; has\_wing\_color: yellow; has\_wing\_color: grey; has\_forehead\_color: black; has\_wing\_pattern: solid; has\_tail\_pattern: solid; has\_tail\_shape: notched\_tail; has\_leg\_color: black; has\_wing\_color: white \\
        Yellow bellied Flycatcher & 14 & has\_forehead\_color: black; has\_belly\_pattern: solid; has\_upper\_tail\_color: white; has\_bill\_length: about\_the\_same\_as\_head; has\_forehead\_color: brown; has\_throat\_color: buff; has\_nape\_color: brown; has\_nape\_color: white; has\_primary\_color: white; has\_bill\_color: buff; has\_primary\_color: grey; has\_breast\_color: black; has\_back\_color: yellow; has\_leg\_color: grey \\
        Frigatebird & 7 & has\_leg\_color: black; has\_under\_tail\_color: grey; has\_breast\_color: white; has\_belly\_pattern: solid; has\_wing\_pattern: multi-colored; has\_bill\_color: black; has\_wing\_color: brown \\
        Northern Fulmar & 1 & has\_bill\_shape: hooked\_seabird \\
        Gadwall & 2 & has\_shape: duck-like; has\_wing\_color: black \\
        American Goldfinch & 14 & has\_back\_color: yellow; has\_upperparts\_color: black; has\_bill\_shape: cone; has\_breast\_color: black; has\_wing\_color: yellow; has\_bill\_color: grey; has\_belly\_pattern: solid; has\_leg\_color: buff; has\_underparts\_color: white; has\_back\_color: brown; has\_back\_color: white; has\_tail\_shape: notched\_tail; has\_under\_tail\_color: white; has\_throat\_color: yellow \\
        European Goldfinch & 11 & has\_forehead\_color: grey; has\_tail\_pattern: multi-colored; has\_breast\_pattern: solid; has\_bill\_shape: all-purpose; has\_bill\_shape: cone; has\_wing\_color: yellow; has\_belly\_color: white; has\_upperparts\_color: yellow; has\_underparts\_color: buff; has\_crown\_color: black; has\_belly\_pattern: solid \\
        Boat tailed Grackle & 14 & has\_tail\_pattern: solid; has\_upperparts\_color: white; has\_wing\_pattern: spotted; has\_forehead\_color: yellow; has\_wing\_color: buff; has\_upperparts\_color: grey; has\_back\_color: grey; has\_breast\_color: grey; has\_belly\_color: brown; has\_belly\_color: white; has\_back\_pattern: multi-colored; has\_bill\_color: buff; has\_breast\_pattern: solid; has\_leg\_color: black \\
        Eared Grebe & 2 & has\_size: small\_(5\_-\_9\_in); has\_belly\_color: grey \\
Horned Grebe & 4 & has\_wing\_color: white; has\_eye\_color: black; has\_nape\_color: grey; has\_shape: duck-like \\
        Pied billed Grebe & 4 & has\_underparts\_color: grey; has\_nape\_color: grey; has\_forehead\_color: grey; has\_belly\_color: yellow \\
        Western Grebe & 2 & has\_eye\_color: black; has\_under\_tail\_color: white \\
        Blue Grosbeak & 7 & has\_back\_pattern: multi-colored; has\_crown\_color: blue; has\_upperparts\_color: black; has\_nape\_color: yellow; has\_wing\_pattern: solid; has\_underparts\_color: white; has\_nape\_color: black \\
        Evening Grosbeak & 6 & has\_upperparts\_color: yellow; has\_nape\_color: brown; has\_belly\_color: brown; has\_wing\_pattern: multi-colored; has\_throat\_color: white; has\_wing\_color: grey \\
        Pine Grosbeak & 14 & has\_breast\_color: grey; has\_belly\_color: brown; has\_size: medium\_(9\_-\_16\_in); has\_throat\_color: grey; has\_underparts\_color: white; has\_underparts\_color: brown; has\_bill\_color: buff; has\_back\_color: grey; has\_belly\_pattern: solid; has\_primary\_color: grey; has\_upperparts\_color: white; has\_bill\_length: about\_the\_same\_as\_head; has\_belly\_color: grey; has\_wing\_color: white \\
        Rose breasted Grosbeak & 14 & has\_breast\_pattern: multi-colored; has\_belly\_color: yellow; has\_primary\_color: grey; has\_back\_color: grey; has\_forehead\_color: blue; has\_nape\_color: brown; has\_underparts\_color: brown; has\_shape: duck-like; has\_leg\_color: grey; has\_head\_pattern: eyebrow; has\_wing\_shape: pointed-wings; has\_throat\_color: buff; has\_wing\_pattern: multi-colored; has\_wing\_pattern: spotted \\
        Pigeon Guillemot & 8 & has\_under\_tail\_color: black; has\_bill\_shape: all-purpose; has\_back\_pattern: solid; has\_wing\_pattern: multi-colored; has\_bill\_shape: hooked\_seabird; has\_tail\_pattern: multi-colored; has\_primary\_color: grey; has\_crown\_color: grey \\
        California Gull & 1 & has\_forehead\_color: white \\
        Glaucous winged Gull & 9 & has\_crown\_color: white; has\_throat\_color: grey; has\_wing\_color: black; has\_bill\_shape: dagger; has\_crown\_color: black; has\_upperparts\_color: black; has\_back\_color: brown; has\_back\_color: yellow; has\_upperparts\_color: grey \\
        Heermann Gull & 3 & has\_primary\_color: grey; has\_wing\_color: brown; has\_bill\_shape: dagger \\
        Ivory Gull & 2 & has\_forehead\_color: white; has\_back\_color: grey \\
        Slaty backed Gull & 3 & has\_crown\_color: white; has\_breast\_pattern: solid; has\_wing\_pattern: solid \\
        Western Gull & 1 & has\_crown\_color: white \\
        Anna Hummingbird & 9 & has\_upperparts\_color: black; has\_breast\_pattern: solid; has\_wing\_shape: pointed-wings; has\_leg\_color: black; has\_bill\_shape: dagger; has\_upper\_tail\_color: white; has\_wing\_color: grey; has\_tail\_pattern: multi-colored; has\_size: small\_(5\_-\_9\_in) \\
        Ruby throated Hummingbird & 19 & has\_underparts\_color: yellow; has\_wing\_color: brown; has\_wing\_color: yellow; has\_bill\_shape: cone; has\_upperparts\_color: brown; has\_upperparts\_color: white; has\_upperparts\_color: buff; has\_underparts\_color: brown; has\_breast\_pattern: striped; has\_back\_color: brown; has\_back\_color: buff; has\_upper\_tail\_color: brown; has\_breast\_color: yellow; has\_breast\_color: white; has\_throat\_color: grey; has\_throat\_color: yellow; has\_size: very\_small\_(3\_-\_5\_in); has\_wing\_color: black; has\_leg\_color: black \\
        Rufous Hummingbird & 4 & has\_bill\_shape: hooked\_seabird; has\_belly\_pattern: solid; has\_upper\_tail\_color: black; has\_back\_pattern: solid \\
        Green Violetear & 8 & has\_wing\_color: white; has\_wing\_color: grey; has\_crown\_color: yellow; has\_wing\_shape: rounded-wings; has\_bill\_length: shorter\_than\_head; has\_size: very\_small\_(3\_-\_5\_in); has\_upperparts\_color: brown; has\_nape\_color: yellow \\
        Blue Jay & 14 & has\_forehead\_color: blue; has\_upper\_tail\_color: buff; has\_breast\_color: black; has\_crown\_color: blue; has\_back\_pattern: striped; has\_bill\_shape: cone; has\_nape\_color: buff; has\_primary\_color: black; has\_leg\_color: buff; has\_wing\_pattern: solid; has\_tail\_pattern: striped; has\_back\_color: grey; has\_eye\_color: black; has\_bill\_color: grey \\
        Florida Jay & 7 & has\_tail\_shape: notched\_tail; has\_crown\_color: grey; has\_crown\_color: blue; has\_wing\_color: white; has\_leg\_color: black; has\_leg\_color: grey; has\_size: small\_(5\_-\_9\_in) \\
Green Jay & 10 & has\_wing\_color: white; has\_head\_pattern: eyebrow; has\_crown\_color: blue; has\_breast\_pattern: multi-colored; has\_under\_tail\_color: buff; has\_primary\_color: grey; has\_wing\_shape: rounded-wings; has\_belly\_pattern: solid; has\_wing\_pattern: multi-colored; has\_primary\_color: yellow \\
        Dark eyed Junco & 12 & has\_throat\_color: grey; has\_breast\_color: white; has\_belly\_color: grey; has\_primary\_color: grey; has\_belly\_pattern: solid; has\_bill\_length: about\_the\_same\_as\_head; has\_belly\_color: white; has\_upper\_tail\_color: grey; has\_wing\_color: black; has\_wing\_shape: rounded-wings; has\_size: small\_(5\_-\_9\_in); has\_primary\_color: yellow \\
        Tropical Kingbird & 12 & has\_crown\_color: grey; has\_underparts\_color: grey; has\_throat\_color: white; has\_breast\_pattern: striped; has\_under\_tail\_color: buff; has\_size: very\_small\_(3\_-\_5\_in); has\_primary\_color: buff; has\_crown\_color: brown; has\_upperparts\_color: black; has\_wing\_pattern: multi-colored; has\_bill\_shape: all-purpose; has\_nape\_color: grey \\
        Gray Kingbird & 12 & has\_nape\_color: grey; has\_wing\_color: yellow; has\_throat\_color: buff; has\_forehead\_color: blue; has\_wing\_color: buff; has\_nape\_color: brown; has\_nape\_color: buff; has\_size: very\_small\_(3\_-\_5\_in); has\_forehead\_color: grey; has\_bill\_shape: hooked\_seabird; has\_back\_color: black; has\_breast\_color: white \\
        Green Kingfisher & 9 & has\_upperparts\_color: white; has\_bill\_shape: dagger; has\_wing\_shape: rounded-wings; has\_breast\_color: white; has\_belly\_color: white; has\_nape\_color: grey; has\_belly\_pattern: solid; has\_wing\_shape: pointed-wings; has\_forehead\_color: black \\
        Pied Kingfisher & 3 & has\_bill\_shape: dagger; has\_breast\_pattern: multi-colored; has\_wing\_pattern: spotted \\
        White breasted Kingfisher & 11 & has\_bill\_shape: hooked\_seabird; has\_under\_tail\_color: buff; has\_upperparts\_color: black; has\_wing\_pattern: multi-colored; has\_wing\_color: yellow; has\_wing\_color: black; has\_head\_pattern: plain; has\_wing\_color: brown; has\_underparts\_color: brown; has\_forehead\_color: brown; has\_bill\_shape: dagger \\
        Red legged Kittiwake & 6 & has\_forehead\_color: white; has\_forehead\_color: grey; has\_bill\_shape: hooked\_seabird; has\_bill\_shape: all-purpose; has\_wing\_color: yellow; has\_wing\_pattern: striped \\
        Horned Lark & 15 & has\_upperparts\_color: buff; has\_back\_pattern: striped; has\_wing\_shape: rounded-wings; has\_throat\_color: yellow; has\_nape\_color: buff; has\_underparts\_color: brown; has\_bill\_shape: all-purpose; has\_underparts\_color: white; has\_primary\_color: white; has\_back\_color: yellow; has\_primary\_color: yellow; has\_wing\_color: grey; has\_underparts\_color: black; has\_bill\_color: black; has\_breast\_pattern: multi-colored \\
        Pacific Loon & 10 & has\_wing\_pattern: spotted; has\_belly\_color: grey; has\_throat\_color: grey; has\_underparts\_color: buff; has\_belly\_color: brown; has\_bill\_length: shorter\_than\_head; has\_back\_pattern: multi-colored; has\_bill\_color: grey; has\_shape: perching-like; has\_nape\_color: grey \\
        Mallard & 6 & has\_back\_pattern: multi-colored; has\_wing\_color: grey; has\_shape: duck-like; has\_wing\_color: black; has\_head\_pattern: plain; has\_wing\_color: brown \\
        Hooded Merganser & 12 & has\_primary\_color: black; has\_eye\_color: black; has\_leg\_color: black; has\_belly\_color: black; has\_wing\_color: brown; has\_underparts\_color: black; has\_back\_color: grey; has\_wing\_shape: pointed-wings; has\_upperparts\_color: black; has\_breast\_color: black; has\_bill\_length: about\_the\_same\_as\_head; has\_wing\_pattern: solid \\
        Red breasted Merganser & 4 & has\_size: medium\_(9\_-\_16\_in); has\_size: small\_(5\_-\_9\_in); has\_back\_pattern: solid; has\_crown\_color: white \\
        Mockingbird & 15 & has\_eye\_color: black; has\_size: very\_small\_(3\_-\_5\_in); has\_underparts\_color: yellow; has\_bill\_shape: all-purpose; has\_breast\_color: yellow; has\_underparts\_color: brown; has\_forehead\_color: blue; has\_belly\_color: brown; has\_shape: duck-like; has\_breast\_pattern: multi-colored; has\_wing\_color: white; has\_wing\_shape: pointed-wings; has\_upperparts\_color: grey; has\_forehead\_color: grey; has\_size: medium\_(9\_-\_16\_in) \\
        Nighthawk & 12 & has\_wing\_color: grey; has\_upperparts\_color: black; has\_upperparts\_color: white; has\_underparts\_color: white; has\_bill\_shape: cone; has\_tail\_shape: notched\_tail; has\_wing\_pattern: spotted; has\_wing\_pattern: multi-colored; has\_back\_pattern: solid; has\_under\_tail\_color: brown; has\_wing\_color: buff; has\_throat\_color: white \\
        Clark Nutcracker & 20 & has\_nape\_color: grey; has\_under\_tail\_color: black; has\_breast\_color: black; has\_underparts\_color: black; has\_bill\_color: grey; has\_underparts\_color: grey; has\_belly\_color: yellow; has\_tail\_shape: notched\_tail; has\_underparts\_color: white; has\_under\_tail\_color: white; has\_wing\_shape: pointed-wings; has\_size: very\_small\_(3\_-\_5\_in); has\_under\_tail\_color: buff; has\_leg\_color: buff; has\_bill\_color: buff; has\_nape\_color: buff; has\_leg\_color: grey; has\_under\_tail\_color: brown; has\_primary\_color: buff; has\_wing\_shape: rounded-wings \\
        White breasted Nuthatch & 11 & has\_throat\_color: grey; has\_wing\_pattern: spotted; has\_crown\_color: yellow; has\_bill\_shape: cone; has\_upperparts\_color: black; has\_shape: duck-like; has\_head\_pattern: plain; has\_wing\_shape: rounded-wings; has\_bill\_color: grey; has\_nape\_color: white; has\_wing\_pattern: solid \\
        Baltimore Oriole & 13 & has\_leg\_color: grey; has\_crown\_color: grey; has\_breast\_color: white; has\_leg\_color: black; has\_wing\_color: brown; has\_underparts\_color: white; has\_back\_color: white; has\_wing\_pattern: striped; has\_upperparts\_color: buff; has\_wing\_color: black; has\_upperparts\_color: white; has\_crown\_color: yellow; has\_primary\_color: yellow \\
        Orchard Oriole & 19 & has\_wing\_color: black; has\_wing\_shape: pointed-wings; has\_eye\_color: black; has\_upperparts\_color: brown; has\_underparts\_color: buff; has\_back\_color: brown; has\_back\_color: white; has\_throat\_color: grey; has\_forehead\_color: brown; has\_nape\_color: buff; has\_belly\_color: grey; has\_belly\_color: white; has\_throat\_color: black; has\_breast\_color: black; has\_shape: perching-like; has\_leg\_color: black; has\_back\_color: black; has\_wing\_pattern: multi-colored; has\_wing\_color: grey \\
        Scott Oriole & 3 & has\_breast\_pattern: multi-colored; has\_throat\_color: black; has\_primary\_color: brown \\
Ovenbird & 10 & has\_bill\_shape: hooked\_seabird; has\_eye\_color: black; has\_leg\_color: grey; has\_back\_pattern: multi-colored; has\_tail\_pattern: striped; has\_upperparts\_color: grey; has\_upperparts\_color: black; has\_breast\_color: buff; has\_head\_pattern: eyebrow; has\_size: small\_(5\_-\_9\_in) \\
        White Pelican & 5 & has\_forehead\_color: white; has\_back\_color: grey; has\_bill\_length: about\_the\_same\_as\_head; has\_back\_color: white; has\_wing\_color: black \\
        Sayornis & 16 & has\_breast\_pattern: multi-colored; has\_crown\_color: yellow; has\_back\_pattern: multi-colored; has\_bill\_color: buff; has\_belly\_color: white; has\_crown\_color: blue; has\_underparts\_color: brown; has\_under\_tail\_color: white; has\_back\_color: white; has\_forehead\_color: blue; has\_nape\_color: yellow; has\_crown\_color: grey; has\_nape\_color: white; has\_size: small\_(5\_-\_9\_in); has\_leg\_color: black; has\_forehead\_color: black \\
        Whip poor Will & 6 & has\_belly\_color: brown; has\_throat\_color: buff; has\_bill\_color: black; has\_shape: duck-like; has\_underparts\_color: brown; has\_leg\_color: buff \\
        Horned Puffin & 16 & has\_forehead\_color: black; has\_leg\_color: black; has\_breast\_pattern: multi-colored; has\_bill\_shape: cone; has\_wing\_pattern: spotted; has\_nape\_color: brown; has\_upperparts\_color: brown; has\_wing\_color: buff; has\_underparts\_color: brown; has\_breast\_color: grey; has\_back\_pattern: solid; has\_wing\_shape: rounded-wings; has\_bill\_length: about\_the\_same\_as\_head; has\_bill\_shape: hooked\_seabird; has\_wing\_pattern: solid; has\_underparts\_color: white \\
        White necked Raven & 5 & has\_underparts\_color: black; has\_upperparts\_color: white; has\_wing\_color: brown; has\_belly\_color: black; has\_nape\_color: white \\
        American Redstart & 10 & has\_breast\_pattern: multi-colored; has\_back\_color: white; has\_wing\_color: black; has\_belly\_pattern: solid; has\_throat\_color: grey; has\_underparts\_color: brown; has\_belly\_color: brown; has\_bill\_shape: hooked\_seabird; has\_upperparts\_color: white; has\_size: small\_(5\_-\_9\_in) \\
        Geococcyx & 10 & has\_wing\_pattern: spotted; has\_underparts\_color: white; has\_breast\_pattern: solid; has\_shape: perching-like; has\_size: small\_(5\_-\_9\_in); has\_underparts\_color: buff; has\_bill\_color: grey; has\_forehead\_color: black; has\_belly\_pattern: solid; has\_bill\_length: about\_the\_same\_as\_head \\
        Loggerhead Shrike & 7 & has\_breast\_color: grey; has\_forehead\_color: grey; has\_bill\_shape: all-purpose; has\_back\_color: grey; has\_underparts\_color: grey; has\_primary\_color: grey; has\_bill\_color: black \\
        Great Grey Shrike & 24 & has\_crown\_color: yellow; has\_bill\_shape: hooked\_seabird; has\_back\_color: grey; has\_belly\_color: grey; has\_size: very\_small\_(3\_-\_5\_in); has\_forehead\_color: blue; has\_belly\_color: buff; has\_eye\_color: black; has\_nape\_color: yellow; has\_wing\_shape: rounded-wings; has\_wing\_color: yellow; has\_belly\_pattern: solid; has\_head\_pattern: plain; has\_wing\_pattern: multi-colored; has\_leg\_color: black; has\_wing\_color: brown; has\_wing\_color: buff; has\_primary\_color: black; has\_breast\_pattern: striped; has\_upperparts\_color: brown; has\_head\_pattern: eyebrow; has\_breast\_color: buff; has\_upperparts\_color: yellow; has\_breast\_color: white \\
        Baird Sparrow & 6 & has\_bill\_color: buff; has\_size: small\_(5\_-\_9\_in); has\_bill\_shape: all-purpose; has\_bill\_length: shorter\_than\_head; has\_size: very\_small\_(3\_-\_5\_in); has\_nape\_color: yellow \\
        Black throated Sparrow & 3 & has\_underparts\_color: grey; has\_forehead\_color: grey; has\_upperparts\_color: yellow \\
        Brewer Sparrow & 3 & has\_belly\_color: buff; has\_breast\_pattern: solid; has\_back\_color: buff \\
        Chipping Sparrow & 9 & has\_upper\_tail\_color: brown; has\_back\_pattern: striped; has\_forehead\_color: brown; has\_breast\_pattern: solid; has\_bill\_color: buff; has\_underparts\_color: brown; has\_leg\_color: buff; has\_tail\_pattern: solid; has\_primary\_color: black \\
        Clay colored Sparrow & 6 & has\_bill\_color: buff; has\_upper\_tail\_color: white; has\_bill\_length: about\_the\_same\_as\_head; has\_wing\_color: grey; has\_belly\_color: brown; has\_back\_pattern: striped \\
        House Sparrow & 10 & has\_leg\_color: buff; has\_belly\_color: buff; has\_crown\_color: grey; has\_breast\_color: grey; has\_tail\_shape: notched\_tail; has\_back\_color: grey; has\_under\_tail\_color: grey; has\_primary\_color: grey; has\_underparts\_color: buff; has\_wing\_color: brown \\
        Field Sparrow & 3 & has\_under\_tail\_color: buff; has\_breast\_pattern: solid; has\_underparts\_color: yellow \\
        Fox Sparrow & 4 & has\_upper\_tail\_color: brown; has\_underparts\_color: buff; has\_underparts\_color: brown; has\_back\_color: buff \\
        Grasshopper Sparrow & 2 & has\_bill\_color: buff; has\_upper\_tail\_color: buff \\
        Harris Sparrow & 13 & has\_breast\_pattern: striped; has\_back\_pattern: striped; has\_throat\_color: black; has\_underparts\_color: white; has\_upper\_tail\_color: black; has\_wing\_shape: rounded-wings; has\_leg\_color: black; has\_bill\_shape: hooked\_seabird; has\_wing\_color: yellow; has\_upperparts\_color: yellow; has\_bill\_length: shorter\_than\_head; has\_back\_color: grey; has\_bill\_shape: all-purpose \\
Henslow Sparrow & 6 & has\_leg\_color: grey; has\_under\_tail\_color: grey; has\_shape: duck-like; has\_breast\_pattern: solid; has\_nape\_color: black; has\_upper\_tail\_color: buff \\
        Le Conte Sparrow & 4 & has\_back\_color: black; has\_nape\_color: brown; has\_bill\_color: black; has\_upperparts\_color: buff \\
        Lincoln Sparrow & 3 & has\_upper\_tail\_color: buff; has\_bill\_color: buff; has\_wing\_color: white \\
        Nelson Sharp tailed Sparrow & 13 & has\_breast\_color: buff; has\_belly\_color: buff; has\_shape: duck-like; has\_wing\_shape: pointed-wings; has\_tail\_pattern: solid; has\_primary\_color: grey; has\_wing\_pattern: solid; has\_belly\_color: brown; has\_leg\_color: buff; has\_back\_color: grey; has\_breast\_pattern: striped; has\_belly\_color: white; has\_under\_tail\_color: brown \\
        Savannah Sparrow & 4 & has\_back\_pattern: striped; has\_upper\_tail\_color: buff; has\_breast\_pattern: solid; has\_bill\_shape: cone \\
        Seaside Sparrow & 11 & has\_eye\_color: black; has\_wing\_color: grey; has\_upperparts\_color: grey; has\_underparts\_color: white; has\_breast\_pattern: multi-colored; has\_upper\_tail\_color: grey; has\_breast\_color: black; has\_forehead\_color: white; has\_under\_tail\_color: white; has\_bill\_color: grey; has\_nape\_color: grey \\
        Song Sparrow & 4 & has\_primary\_color: brown; has\_wing\_shape: pointed-wings; has\_primary\_color: buff; has\_breast\_color: buff \\
        Tree Sparrow & 12 & has\_back\_pattern: striped; has\_size: very\_small\_(3\_-\_5\_in); has\_breast\_color: black; has\_forehead\_color: blue; has\_belly\_color: grey; has\_wing\_color: black; has\_throat\_color: yellow; has\_underparts\_color: grey; has\_forehead\_color: grey; has\_forehead\_color: black; has\_breast\_pattern: solid; has\_leg\_color: buff \\
        Vesper Sparrow & 2 & has\_bill\_color: buff; has\_underparts\_color: buff \\
        White crowned Sparrow & 3 & has\_throat\_color: grey; has\_under\_tail\_color: grey; has\_breast\_color: buff \\
        White throated Sparrow & 3 & has\_throat\_color: grey; has\_under\_tail\_color: brown; has\_forehead\_color: black \\
        Cape Glossy Starling & 6 & has\_tail\_pattern: solid; has\_underparts\_color: black; has\_forehead\_color: blue; has\_head\_pattern: plain; has\_bill\_shape: cone; has\_back\_pattern: multi-colored \\
        Bank Swallow & 14 & has\_size: very\_small\_(3\_-\_5\_in); has\_throat\_color: grey; has\_belly\_color: brown; has\_wing\_pattern: striped; has\_wing\_shape: rounded-wings; has\_tail\_shape: notched\_tail; has\_primary\_color: yellow; has\_shape: duck-like; has\_nape\_color: black; has\_breast\_color: black; has\_wing\_pattern: spotted; has\_primary\_color: black; has\_under\_tail\_color: brown; has\_back\_color: brown \\
        Barn Swallow & 17 & has\_bill\_color: black; has\_wing\_pattern: striped; has\_forehead\_color: white; has\_crown\_color: white; has\_breast\_color: white; has\_back\_pattern: striped; has\_wing\_color: buff; has\_crown\_color: blue; has\_under\_tail\_color: brown; has\_under\_tail\_color: buff; has\_wing\_color: brown; has\_tail\_pattern: multi-colored; has\_primary\_color: buff; has\_crown\_color: yellow; has\_wing\_color: grey; has\_belly\_color: buff; has\_belly\_pattern: solid \\
        Cliff Swallow & 9 & has\_eye\_color: black; has\_belly\_color: yellow; has\_crown\_color: blue; has\_throat\_color: yellow; has\_primary\_color: yellow; has\_nape\_color: grey; has\_breast\_color: brown; has\_bill\_length: shorter\_than\_head; has\_underparts\_color: white \\
        Scarlet Tanager & 18 & has\_bill\_color: buff; has\_wing\_pattern: spotted; has\_underparts\_color: brown; has\_belly\_color: white; has\_shape: duck-like; has\_nape\_color: white; has\_primary\_color: grey; has\_belly\_pattern: solid; has\_breast\_color: grey; has\_bill\_length: about\_the\_same\_as\_head; has\_belly\_color: brown; has\_upperparts\_color: grey; has\_leg\_color: grey; has\_bill\_color: grey; has\_eye\_color: black; has\_breast\_pattern: solid; has\_tail\_shape: notched\_tail; has\_bill\_shape: all-purpose \\
        Summer Tanager & 18 & has\_crown\_color: blue; has\_breast\_color: white; has\_belly\_color: brown; has\_belly\_color: white; has\_underparts\_color: brown; has\_nape\_color: brown; has\_breast\_color: brown; has\_belly\_pattern: solid; has\_leg\_color: black; has\_wing\_color: grey; has\_upper\_tail\_color: grey; has\_back\_color: grey; has\_belly\_color: grey; has\_bill\_length: shorter\_than\_head; has\_breast\_pattern: solid; has\_bill\_shape: all-purpose; has\_back\_pattern: solid; has\_wing\_color: black \\
        Black Tern & 6 & has\_leg\_color: black; has\_wing\_color: black; has\_upper\_tail\_color: black; has\_belly\_pattern: solid; has\_underparts\_color: black; has\_back\_color: brown \\
        Elegant Tern & 3 & has\_under\_tail\_color: white; has\_wing\_color: black; has\_primary\_color: grey \\
        Least Tern & 4 & has\_nape\_color: white; has\_wing\_color: black; has\_bill\_shape: hooked\_seabird; has\_wing\_color: white \\
Green tailed Towhee & 2 & has\_throat\_color: grey; has\_wing\_color: black \\
        Brown Thrasher & 4 & has\_back\_pattern: striped; has\_belly\_color: white; has\_wing\_shape: rounded-wings; has\_upper\_tail\_color: brown \\
        Sage Thrasher & 8 & has\_shape: duck-like; has\_back\_pattern: multi-colored; has\_breast\_pattern: solid; has\_nape\_color: black; has\_forehead\_color: grey; has\_size: small\_(5\_-\_9\_in); has\_wing\_color: buff; has\_upper\_tail\_color: brown \\
        Black capped Vireo & 14 & has\_breast\_pattern: multi-colored; has\_wing\_pattern: solid; has\_upper\_tail\_color: grey; has\_back\_pattern: multi-colored; has\_bill\_shape: hooked\_seabird; has\_belly\_pattern: solid; has\_belly\_color: grey; has\_wing\_pattern: spotted; has\_wing\_color: brown; has\_upperparts\_color: brown; has\_forehead\_color: blue; has\_forehead\_color: yellow; has\_under\_tail\_color: grey; has\_belly\_color: white \\
        Blue headed Vireo & 9 & has\_bill\_shape: hooked\_seabird; has\_forehead\_color: grey; has\_back\_color: grey; has\_wing\_color: grey; has\_upperparts\_color: grey; has\_throat\_color: yellow; has\_wing\_color: brown; has\_underparts\_color: grey; has\_underparts\_color: buff \\
        Philadelphia Vireo & 4 & has\_crown\_color: grey; has\_back\_pattern: solid; has\_under\_tail\_color: black; has\_tail\_pattern: solid \\
        Red eyed Vireo & 14 & has\_upper\_tail\_color: black; has\_breast\_color: black; has\_throat\_color: buff; has\_underparts\_color: grey; has\_breast\_color: grey; has\_forehead\_color: black; has\_under\_tail\_color: black; has\_nape\_color: brown; has\_nape\_color: black; has\_belly\_color: brown; has\_eye\_color: black; has\_leg\_color: grey; has\_nape\_color: buff; has\_belly\_color: white \\
        Warbling Vireo & 4 & has\_nape\_color: grey; has\_head\_pattern: eyebrow; has\_underparts\_color: white; has\_back\_color: black \\
        White eyed Vireo & 11 & has\_nape\_color: grey; has\_throat\_color: grey; has\_wing\_pattern: striped; has\_primary\_color: grey; has\_bill\_shape: cone; has\_bill\_length: about\_the\_same\_as\_head; has\_tail\_pattern: multi-colored; has\_wing\_color: grey; has\_head\_pattern: plain; has\_leg\_color: grey; has\_leg\_color: black \\
        Black and white Warbler & 4 & has\_upperparts\_color: white; has\_back\_pattern: striped; has\_upperparts\_color: black; has\_upper\_tail\_color: buff \\
        Black throated Blue Warbler & 16 & has\_breast\_pattern: multi-colored; has\_bill\_shape: hooked\_seabird; has\_belly\_color: buff; has\_back\_pattern: multi-colored; has\_bill\_color: buff; has\_upperparts\_color: grey; has\_wing\_shape: pointed-wings; has\_nape\_color: black; has\_throat\_color: black; has\_breast\_color: white; has\_belly\_color: black; has\_breast\_pattern: solid; has\_size: very\_small\_(3\_-\_5\_in); has\_wing\_color: brown; has\_nape\_color: white; has\_wing\_pattern: multi-colored \\
        Blue winged Warbler & 6 & has\_breast\_color: brown; has\_under\_tail\_color: buff; has\_throat\_color: buff; has\_wing\_color: yellow; has\_head\_pattern: plain; has\_crown\_color: yellow \\
        Canada Warbler & 7 & has\_back\_pattern: multi-colored; has\_bill\_shape: all-purpose; has\_wing\_shape: pointed-wings; has\_throat\_color: grey; has\_upperparts\_color: black; has\_breast\_pattern: multi-colored; has\_under\_tail\_color: grey \\
        Cerulean Warbler & 4 & has\_forehead\_color: blue; has\_underparts\_color: white; has\_size: very\_small\_(3\_-\_5\_in); has\_bill\_shape: cone \\
        Chestnut sided Warbler & 13 & has\_breast\_pattern: multi-colored; has\_breast\_color: grey; has\_throat\_color: buff; has\_wing\_pattern: spotted; has\_shape: duck-like; has\_tail\_pattern: striped; has\_belly\_color: buff; has\_underparts\_color: brown; has\_breast\_color: brown; has\_crown\_color: white; has\_crown\_color: grey; has\_bill\_color: buff; has\_eye\_color: black \\
        Golden winged Warbler & 9 & has\_bill\_color: black; has\_leg\_color: grey; has\_wing\_shape: pointed-wings; has\_back\_color: grey; has\_belly\_color: white; has\_wing\_color: black; has\_wing\_pattern: multi-colored; has\_underparts\_color: white; has\_belly\_color: grey \\
        Hooded Warbler & 8 & has\_breast\_color: yellow; has\_leg\_color: buff; has\_wing\_color: buff; has\_nape\_color: grey; has\_back\_pattern: multi-colored; has\_belly\_color: white; has\_wing\_color: brown; has\_primary\_color: white \\
        Kentucky Warbler & 8 & has\_back\_color: yellow; has\_upperparts\_color: grey; has\_underparts\_color: black; has\_wing\_shape: pointed-wings; has\_upperparts\_color: white; has\_leg\_color: grey; has\_size: small\_(5\_-\_9\_in); has\_forehead\_color: black \\
        Mourning Warbler & 8 & has\_eye\_color: black; has\_underparts\_color: grey; has\_upperparts\_color: black; has\_under\_tail\_color: white; has\_throat\_color: buff; has\_breast\_pattern: multi-colored; has\_throat\_color: grey; has\_primary\_color: grey \\
        Myrtle Warbler & 11 & has\_forehead\_color: brown; has\_breast\_pattern: striped; has\_under\_tail\_color: buff; has\_nape\_color: brown; has\_back\_pattern: multi-colored; has\_wing\_pattern: solid; has\_bill\_shape: dagger; has\_underparts\_color: yellow; has\_bill\_length: shorter\_than\_head; has\_wing\_shape: pointed-wings; has\_leg\_color: black \\
Nashville Warbler & 3 & has\_bill\_color: grey; has\_underparts\_color: buff; has\_forehead\_color: blue \\
        Orange crowned Warbler & 11 & has\_throat\_color: buff; has\_bill\_color: black; has\_bill\_shape: dagger; has\_underparts\_color: grey; has\_upperparts\_color: yellow; has\_back\_color: yellow; has\_wing\_pattern: multi-colored; has\_back\_color: black; has\_back\_color: grey; has\_bill\_shape: all-purpose; has\_primary\_color: black \\
        Palm Warbler & 26 & has\_bill\_shape: dagger; has\_upperparts\_color: white; has\_wing\_color: white; has\_back\_color: black; has\_back\_color: white; has\_upper\_tail\_color: black; has\_wing\_color: black; has\_head\_pattern: eyebrow; has\_eye\_color: black; has\_breast\_pattern: multi-colored; has\_underparts\_color: buff; has\_breast\_color: black; has\_throat\_color: white; has\_under\_tail\_color: black; has\_nape\_color: black; has\_belly\_color: brown; has\_belly\_color: white; has\_tail\_pattern: striped; has\_primary\_color: black; has\_wing\_pattern: spotted; has\_under\_tail\_color: white; has\_size: small\_(5\_-\_9\_in); has\_forehead\_color: blue; has\_wing\_pattern: striped; has\_wing\_pattern: multi-colored; has\_throat\_color: grey \\
        Pine Warbler & 1 & has\_crown\_color: yellow \\
        Prairie Warbler & 4 & has\_crown\_color: yellow; has\_size: medium\_(9\_-\_16\_in); has\_nape\_color: yellow; has\_primary\_color: white \\
        Swainson Warbler & 6 & has\_bill\_shape: hooked\_seabird; has\_tail\_pattern: striped; has\_upperparts\_color: black; has\_wing\_pattern: solid; has\_forehead\_color: brown; has\_breast\_pattern: solid \\
        Tennessee Warbler & 5 & has\_crown\_color: grey; has\_wing\_shape: pointed-wings; has\_bill\_color: black; has\_back\_pattern: multi-colored; has\_crown\_color: brown \\
        Wilson Warbler & 2 & has\_back\_color: yellow; has\_leg\_color: black \\
        Worm eating Warbler & 6 & has\_eye\_color: black; has\_forehead\_color: white; has\_wing\_pattern: solid; has\_upperparts\_color: black; has\_forehead\_color: black; has\_forehead\_color: yellow \\
        Yellow Warbler & 12 & has\_back\_color: yellow; has\_breast\_pattern: solid; has\_bill\_color: grey; has\_primary\_color: yellow; has\_bill\_shape: cone; has\_forehead\_color: grey; has\_wing\_pattern: striped; has\_breast\_color: grey; has\_breast\_color: buff; has\_throat\_color: grey; has\_eye\_color: black; has\_back\_pattern: multi-colored \\
        Northern Waterthrush & 5 & has\_head\_pattern: eyebrow; has\_under\_tail\_color: black; has\_underparts\_color: buff; has\_wing\_color: grey; has\_wing\_shape: pointed-wings \\
        Louisiana Waterthrush & 7 & has\_head\_pattern: eyebrow; has\_breast\_color: black; has\_underparts\_color: buff; has\_tail\_shape: notched\_tail; has\_underparts\_color: brown; has\_wing\_color: grey; has\_tail\_pattern: striped \\
        Bohemian Waxwing & 6 & has\_upperparts\_color: grey; has\_wing\_shape: pointed-wings; has\_belly\_color: grey; has\_wing\_shape: rounded-wings; has\_tail\_pattern: multi-colored; has\_forehead\_color: grey \\
        Cedar Waxwing & 3 & has\_tail\_pattern: multi-colored; has\_nape\_color: buff; has\_upper\_tail\_color: brown \\
        American Three toed Woodpecker & 10 & has\_bill\_shape: dagger; has\_breast\_pattern: solid; has\_leg\_color: grey; has\_head\_pattern: eyebrow; has\_wing\_color: black; has\_back\_pattern: multi-colored; has\_wing\_shape: pointed-wings; has\_tail\_pattern: striped; has\_belly\_pattern: solid; has\_tail\_pattern: solid \\
        Pileated Woodpecker & 17 & has\_under\_tail\_color: black; has\_bill\_color: grey; has\_upper\_tail\_color: black; has\_wing\_shape: pointed-wings; has\_forehead\_color: white; has\_leg\_color: grey; has\_primary\_color: grey; has\_upper\_tail\_color: buff; has\_breast\_color: yellow; has\_throat\_color: yellow; has\_bill\_length: shorter\_than\_head; has\_eye\_color: black; has\_wing\_color: grey; has\_forehead\_color: brown; has\_size: medium\_(9\_-\_16\_in); has\_underparts\_color: white; has\_nape\_color: white \\
        Red bellied Woodpecker & 6 & has\_bill\_shape: dagger; has\_tail\_pattern: striped; has\_wing\_color: brown; has\_back\_pattern: solid; has\_crown\_color: black; has\_wing\_shape: pointed-wings \\
        Red cockaded Woodpecker & 5 & has\_nape\_color: black; has\_head\_pattern: eyebrow; has\_belly\_pattern: solid; has\_wing\_shape: pointed-wings; has\_throat\_color: white \\
        Downy Woodpecker & 25 & has\_breast\_pattern: multi-colored; has\_underparts\_color: yellow; has\_upperparts\_color: brown; has\_upperparts\_color: black; has\_underparts\_color: black; has\_eye\_color: black; has\_back\_pattern: multi-colored; has\_bill\_length: about\_the\_same\_as\_head; has\_primary\_color: white; has\_wing\_color: black; has\_back\_pattern: solid; has\_back\_color: white; has\_forehead\_color: white; has\_size: small\_(5\_-\_9\_in); has\_wing\_shape: pointed-wings; has\_shape: perching-like; has\_bill\_shape: all-purpose; has\_upperparts\_color: white; has\_wing\_pattern: striped; has\_underparts\_color: grey; has\_underparts\_color: buff; has\_bill\_length: shorter\_than\_head; has\_bill\_shape: cone; has\_wing\_color: brown; has\_breast\_pattern: solid \\
        Bewick Wren & 4 & has\_tail\_pattern: striped; has\_wing\_pattern: solid; has\_tail\_shape: notched\_tail; has\_bill\_shape: all-purpose \\
Cactus Wren & 3 & has\_wing\_pattern: spotted; has\_crown\_color: white; has\_head\_pattern: eyebrow \\
        Carolina Wren & 8 & has\_tail\_pattern: striped; has\_underparts\_color: grey; has\_nape\_color: buff; has\_forehead\_color: brown; has\_bill\_color: grey; has\_primary\_color: white; has\_forehead\_color: blue; has\_crown\_color: blue \\
        House Wren & 3 & has\_tail\_pattern: striped; has\_belly\_color: buff; has\_belly\_color: white \\
        Marsh Wren & 10 & has\_tail\_pattern: striped; has\_wing\_pattern: spotted; has\_back\_color: grey; has\_forehead\_color: yellow; has\_back\_pattern: multi-colored; has\_eye\_color: black; has\_wing\_shape: rounded-wings; has\_forehead\_color: brown; has\_breast\_pattern: solid; has\_back\_color: black \\
        Rock Wren & 4 & has\_back\_pattern: striped; has\_crown\_color: brown; has\_belly\_color: brown; has\_under\_tail\_color: black \\
        Winter Wren & 3 & has\_breast\_color: brown; has\_breast\_pattern: striped; has\_size: medium\_(9\_-\_16\_in) \\
\end{longtable}
}